\documentclass[lettersize,journal]{IEEEtran}
\usepackage{amsmath,amsfonts}
\usepackage{algorithmic}
\usepackage{algorithm}
\usepackage{array}
\usepackage[caption=false,font=normalsize,labelfont=sf,textfont=sf]{subfig}
\usepackage{textcomp}
\usepackage{stfloats}
\usepackage{url}
\usepackage{verbatim}
\usepackage{graphicx}
\usepackage{cite}
\usepackage{booktabs}
\begin{document}

\def\mtd{4DLidarOpen}

\newcommand*\samethanks[1][\value{footnote}]{\footnotemark[#1]}
\newcommand*\samethanksrepeat[1][\value{footnote}]{\footnotemark[#1]}

\title{\mtd{}: An Open 4D FMCW Lidar Dataset for Motion-Aware Autonomous Driving}

\author{Kane Qian\textsuperscript{1}, Xin Zhao\textsuperscript{2}, Yining Shi\textsuperscript{1}, Rujun Yan\textsuperscript{1}, Zhengqing Pan\textsuperscript{2}, Kaojin Zhu\textsuperscript{2}, Mengmeng Yang\textsuperscript{1}, Kai Sun\textsuperscript{2}, Diange Yang\textsuperscript{1}, Kun Jiang\textsuperscript{1,\textdagger}
\thanks{\textsuperscript{1}Tsinghua University.}
\thanks{\textsuperscript{2}Hesai Technology Co., Ltd.}
\thanks{\textsuperscript{\textdagger
}Corresponding author: Kun Jiang.}
\thanks{This work was supported in part by the National Natural Science Foundation of China (52372414, 52394264, 52472449).}
}

\markboth{Submitted to IEEE Transactions on Instrumentation and Measurement}%
{Shell \MakeLowercase{\textit{et al.}}: A Sample Article Using IEEEtran.cls for IEEE Journals}


\maketitle

\begin{abstract}
We present 4DLidarOpen, a large-scale open multi-modal dataset for autonomous driving, centered on 4D frequency-modulated continuous-wave (FMCW) Lidar sensing. Unlike conventional time-of-flight Lidar datasets that mainly provide geometric measurements, 4DLidarOpen includes point-wise radial velocity measurements from a forward-facing 4D FMCW Lidar, together with multiple Lidars of different types, including rotating, solid-state, and blind-spot variants, surround-view cameras, and 6-DOF ego-vehicle poses. The dataset was collected in complex urban environments in Beijing and covers dense pedestrian interactions, congested traffic, high-speed driving, and unprotected maneuvers.

4DLidarOpen provides synchronized multi-sensor data and 3D bounding-box annotations with persistent track IDs across five object categories. A hybrid annotation strategy is adopted, where large-scale auto-labeled data support scalable training and human experts refine annotations for the human-annotated training and validation sets. Based on this dataset, we establish benchmarks for 3D object detection, bird’s-eye view (BEV) segmentation and flow prediction, and motion forecasting with planning.

Extensive experiments show that direct velocity measurements from 4D FMCW Lidar provide complementary motion cues for dynamic-scene understanding. Compared with geometric-only sensing, the velocity-aware representation improves motion-related perception and downstream forecasting and planning, especially in scenarios involving vulnerable road users and fast-moving objects. These results indicate that 4D FMCW Lidar is a promising sensing modality for motion-aware autonomous driving. The dataset and evaluation toolkit are publicly released to support research on 4D scene understanding, multi-Lidar fusion, and velocity-aware perception and planning.
\end{abstract}
\section{Introduction}

As autonomous driving systems progress toward Level 3 autonomy and beyond, end-to-end (E2E) learning has become an increasingly important paradigm for integrating perception, prediction, and planning \cite{chen2024end, jiang2025survey}. Unlike traditional modular pipelines that separate perception, prediction, and planning into distinct components, E2E approaches learn a direct mapping from raw sensor inputs to control commands. This integration enables global optimization and fosters behaviorally consistent agents that can better handle complex traffic scenarios. Recent advances in world models and unified driving frameworks further suggest that autonomous systems will depend not just on increasingly large policy networks, but on expressive, motion-aware scene representations that capture the dynamic nature of the driving environment \cite{qian2025agentthink, yu20264d}. However, despite these advances, contemporary datasets often fall short of providing the temporal resolution and motion fidelity required for reliable 4D scene understanding.

\begin{figure}[htbp]
    \centering
    \includegraphics[width=0.45\textwidth]{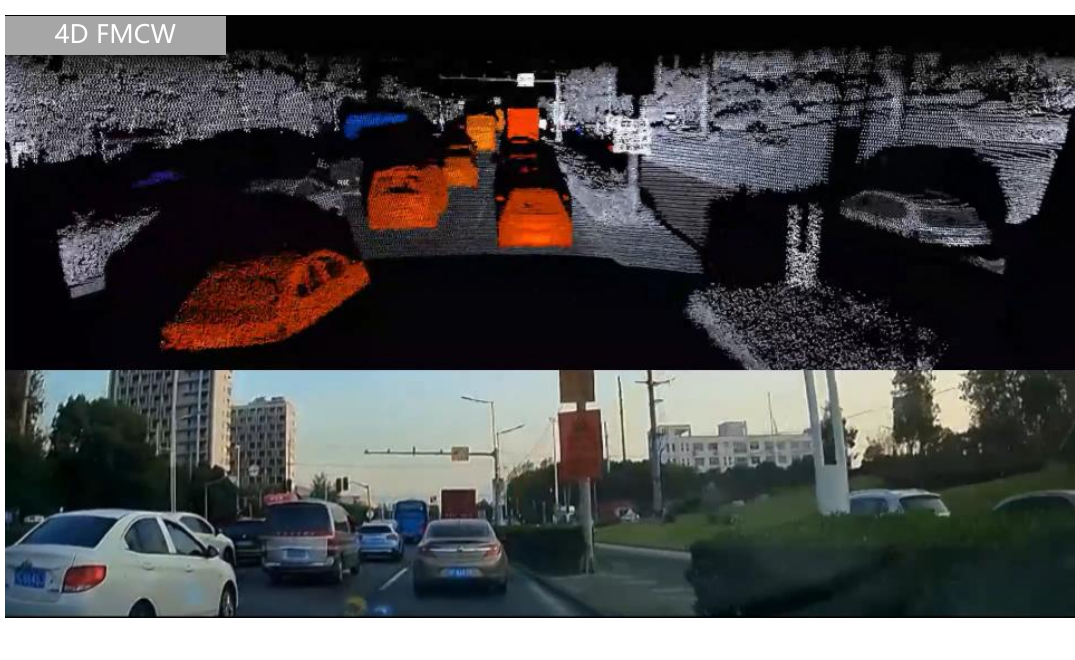}
    \caption{4D FMCW sample showing raw point cloud data with point-wise radial velocity information for motion-aware scene analysis.}
    \label{fig:data-sample-wt}
\end{figure}

At its core, autonomous driving requires an intelligent agent to construct a temporally coherent representation of a dynamic 3D world. This task extends far beyond simple semantic recognition, as the agent must simultaneously estimate 3D geometry, track motion, maintain temporal continuity, reason about interactions, and infer interaction patterns, all under strict real-time constraints. The fundamental challenge involves answering critical questions about the scene: identifying the objects present, determining their locations, understanding their motion patterns, predicting their interactions with the ego vehicle, and anticipating how the scene will evolve in the coming seconds. Consequently, autonomous driving is inherently a 4D scene-understanding problem rather than a static scene-parsing task \cite{shi2024streamingflow}.

Autonomous driving research is commonly organized around two representative paradigms. Modular systems decompose the driving pipeline into separate perception, tracking, prediction, planning, and control components, which offers interpretable intermediate representations and engineering flexibility. In contrast, E2E approaches learn shared latent representations, such as bird’s-eye view (BEV) grids, occupancy volumes, or joint prediction-planning embeddings \cite{chen2024vad}, integrating these stages into a unified framework. Although both paradigms have pushed the state of the art forward, their scene representations still rely heavily on object categories, geometric localization, and short-horizon temporal aggregation. This limits their ability to capture the full dynamics of real-world traffic.

These limitations become particularly pronounced in complex real-world scenarios. Frame-centric perception methods often dilute temporal consistency and obscure continuous motion cues, while indirect motion inference through techniques like frame-to-frame association or heuristic feature aggregation degrades performance in long-range, high-speed, or densely interactive situations. Furthermore, scene understanding approaches that focus solely on semantics and geometry struggle to fully characterize dynamic traffic evolution, which is essential for accurate motion forecasting, interaction reasoning, and effective planning. Compounding these challenges, most existing public datasets are built around conventional camera and time-of-flight Lidar setups that provide strong geometric measurements but limited direct observability of object motion. Consequently, many downstream tasks must reconstruct dynamics indirectly, introducing ambiguity and reducing robustness in safety-critical situations.

Recent progress in autonomous driving world models underscores this limitation. DriveWorld demonstrates that autonomous driving should be approached as a 4D scene understanding problem, highlighting the value of explicitly learning spatiotemporal representations for perception, forecasting, occupancy prediction, and planning \cite{min2024driveworld}. Similarly, World4Drive shows that latent world modeling can support end-to-end planning by capturing intention-aware physical scene evolution \cite{zheng2025world4drive}. These developments suggest that 4D understanding is not merely an auxiliary capability but a key foundation for motion-aware autonomous driving. This mirrors the human perception process, where the brain seamlessly integrates motion cues, such as the speed of surrounding objects, to anticipate and react to changing situations. For an autonomous agent, directly perceiving motion states rather than inferring them indirectly is critical for building a robust and predictive understanding of the dynamic world.

Motivated by these observations, we introduce 4DLidarOpen, a large-scale multi-modal dataset that bridges 4D scene-understanding research and heterogeneous Lidar sensing. The dataset provides synchronized surround-view cameras, multi-Lidar streams, and 6-DOF ego-vehicle poses, forming a comprehensive sensing suite for autonomous driving research. 4DLidarOpen includes a diverse Lidar configuration:
\begin{enumerate}
    \item A 4D FMCW Lidar that delivers (x,y,z,v) point clouds through forward-facing scans, providing instantaneous radial velocity for each point.
    \item A 360\textdegree{} rotating OT Lidar with 300-meter range, serving as the primary reference for high-quality annotations.
    \item A solid-state AT Lidar that provides high-density forward-facing scans, ideal for ADAS applications.
    \item Two ATX blind-spot Lidars that monitor near-field regions, capturing objects in areas often missed by other sensors.
\end{enumerate}

All sensors in 4DLidarOpen are hardware-synchronized and jointly calibrated with high-precision extrinsic calibration \cite{Geiger13ijrr_KITTIVisionMeetsRobotics}, ensuring coherent multi-modal data fusion. Beyond raw sensor streams, 4DLidarOpen provides detailed per-sensor annotations, including 3D bounding boxes with persistent tracklets that maintain object identity across frames \cite{Caesar20cvpr_NuScenes,Geiger12cvpr_KITTIAreWeReady}.

Leveraging this rich dataset, we address a critical research question: how does 4D FMCW Lidar sensing enhance autonomous driving capabilities through richer 4D scene understanding? To answer this, we conduct systematic benchmarks that compare the performance of forward-facing AT, 4D FMCW, and 360\textdegree{} OT Lidar configurations across a comprehensive suite of tasks. These benchmarks span 3D object detection \cite{Lang19cvpr_PointPillars,Yin21cvpr_CenterPoint}, BEV segmentation and flow prediction \cite{Behley19iccv_SemanticKITTI,Liu22arxiv_BEVFusion}, and motion forecasting coupled with path planning \cite{Liang20eccv_LaneGCN,Mercat20icra_MultiheadAttentionForecasting}. In doing so, 4DLidarOpen provides a concrete testbed for studying how motion-aware sensing benefits object-level perception, scene-level understanding, and downstream driving tasks.

As illustrated in Figure~\ref{fig:data-sample-wt}, the 4D FMCW Lidar captures instantaneous radial velocity information for each point, providing significant advantages for motion analysis and dynamic object detection. This velocity-aware point cloud representation enables more accurate tracking of moving objects such as pedestrians, cyclists, and vehicles, as the radial velocity can be directly measured without requiring frame-to-frame differentiation.

4DLidarOpen makes three principal contributions to the autonomous driving research community:
\begin{enumerate}
    \item We release 4DLidarOpen, an open multi-modal autonomous driving dataset centered on 4D FMCW Lidar sensing. The dataset integrates a forward-facing 4D FMCW Lidar, rotating OT Lidar, solid-state AT Lidar, ATX blind-spot Lidars, synchronized surround-view cameras, and 6-DOF ego-vehicle poses.

    \item We provide unified benchmarks for 3D object detection, BEV segmentation and flow prediction, and motion forecasting with planning, enabling evaluation from object-level perception to downstream driving tasks.

    \item We empirically quantify the role of point-wise radial velocity measurements and show that 4D FMCW Lidar provides complementary motion cues for dynamic-scene understanding and downstream forecasting and planning.
\end{enumerate}

4DLidarOpen and its open-source evaluation toolkit are publicly available at \url{https://github.com/haopen-dataset/haopen}, providing the research community with a reproducible foundation for advancing 4D scene-understanding and motion-aware driving technologies. By making this resource available, we aim to support reproducible research on motion-aware perception, prediction, and planning in dynamic traffic environments.
\section{Related Work}

\subsection{Sensor Datasets} 

Over the past decade, the autonomous driving research community has seen the release of numerous sensor-rich datasets designed to train and benchmark perception, prediction, and planning modules. Pioneering datasets like KITTI\cite{Geiger12cvpr_KITTIAreWeReady} and Cityscapes\cite{cordtsCityscapesDatasetSemantic2016} laid the foundation for calibrated multi-modal data collection, yet their relatively limited 15 hours of driving time and lack of behavioral labels make them unsuitable for end-to-end learning approaches. Building on this foundation, nuScenes\cite{Caesar20cvpr_NuScenes} introduced 1,000 twenty-second clips with 1.4 million 3D object boxes, while ApolloScape\cite{ApolloScape} added 140,000 high-resolution frames with pixel-level semantics and lane masks. Waymo Open\cite{Sun20cvpr_WaymoOpenDataset} further expanded the scope with 1,950 scenes recorded across urban, suburban, and highway domains using five Lidars and five cameras. More recently, datasets such as CODA\cite{li2022coda} and PandaSet\cite{xiao2021pandaset} have focused on capturing rare events. However, few public datasets provide both surround-view geometric sensing and point-wise instantaneous velocity measurements, which limits systematic studies of motion-aware 4D scene understanding.

Three key gaps persist in the current landscape of autonomous driving datasets. First, most existing datasets rely on homogeneous sensor setups—typically frontal 120° cameras and a single rooftop Lidar—while modern production vehicles increasingly employ surround-view camera rigs, side and rear radars, and multiple solid-state Lidars. Second, the Lidar specifications in existing datasets often lag behind the capabilities of contemporary assisted-driving fleets, limiting their relevance for next-generation systems. Third, and most critically, the perceptual value of 4D FMCW Lidar—with its ability to provide instantaneous radial velocity measurements—remains largely unexplored in public benchmarks, despite its potential value for dynamic scene understanding.

While simulators such as CARLA\cite{dosovitskiy2017carla} help address data scarcity, the inherent sim-to-real gap limits their utility for safety-critical deployment. This motivates large-scale, geographically diverse, and action-labeled real-world datasets that support the joint study of perception and control, which is one of the goals of 4DLidarOpen.

\subsection{Recent Advances in End-to-End Driving}
End-to-end (E2E) driving represents a paradigm shift in autonomous vehicle development, learning a direct mapping from raw sensor inputs to trajectories or control commands without relying on hand-engineered intermediates like 3D bounding boxes, motion priors \cite{qian2025lego}, or HD maps~\cite{paden2016survey}. Bojarski et al.~\cite{bojarski2016end} popularized this approach by training a convolutional neural network (CNN) to regress steering angles from monocular video. Despite its simplicity, their system achieved robust highway lane keeping, establishing imitation learning (IL) as a foundational approach for E2E driving~\cite{codevilla2018end, xu2017end}. Subsequent research has expanded this paradigm to include reinforcement learning (RL)~\cite{kendall2019learning, schulman2017proximal}, inverse reinforcement learning, and behavior cloning with data augmentation techniques.

Progress in E2E driving has unfolded along three primary axes: input modalities, policy architectures, and training objectives. Inputs have evolved from monocular RGB~\cite{bojarski2016end} to encompass surround-view camera rigs, Lidar point clouds~\cite{liu2023bevfusion}, and heterogeneous sensor fusion~\cite{prakash2021transfuser, chitta2022transfuser}. TransFuser\cite{chitta2022transfuser} leverages transformers to fuse Lidar and camera features, while BEVFusion\cite{liu2023bevfusion} unifies multi-sensor cues in a bird’s-eye view (BEV) grid. Recent vectorized approaches like VAD~\cite{jiang2023vad} efficiently encode scene information for planning tasks.

Architectures have matured from simple CNNs to sophisticated spatio-temporal networks. Recurrent neural networks and temporal convolutions encode historical information~\cite{altche2017lstm}, while attention mechanisms dynamically weight scene entities based on their relevance. Multi-task frameworks jointly regress trajectories, segmentation, and occupancy, using auxiliary supervision to improve generalization~\cite{crawshaw2020multi, ishihara2021multi}. UniAD~\cite{hu2023uniad} represents a significant advance by unifying perception, prediction, and planning in a single framework, while recent generative world models~\cite{bartoccioni2025vavim} and diffusion models~\cite{peebles2023scalable, diffusiondrive} synthesize diverse scenarios to enhance planning capabilities.

Training objectives have diversified beyond pure imitation learning. Reinforcement learning with carefully designed reward functions addresses the distribution shift problem inherent in imitation learning~\cite{kendall2019learning}. Adversarial training improves robustness to distribution shifts~\cite{jaeger2023hidden}, and safety-constrained optimization incorporates collision avoidance as differentiable costs~\cite{zeng2019end, sadat2020perceive}. Direct Preference Optimization (DPO)~\cite{rafailov2023direct} has also been adapted to align driving policies with human preferences, improving the naturalness of generated behaviors.

A particularly significant recent development is the integration of Large Language Models (LLMs)~\cite{touvron2023llama1, touvron2023llama2} and Vision-Language Models (VLMs)~\cite{radford2021learning} to inject commonsense reasoning and interpretability into driving systems. Works like GPT-Driver~\cite{mao2023gpt}, DriveMLM~\cite{wang2023drivemlm}, and LmDrive~\cite{shao2024lmdrive} use LLMs as planners or reasoning engines. Subsequent studies have enhanced this paradigm with agent-based reasoning~\cite{mao2023agentdriver, hou2025driveagent}, knowledge grounding~\cite{wen2023dilu, jiang2024koma}, and visual instruction-tuning~\cite{zhang2024instruct, ishaq2025drivelmm}. The emergence of Vision-Language-Action (VLA) models marks a pivotal shift towards true end-to-end systems that process raw pixels and output actionable control commands. DriveVLM~\cite{huang2024drivemm}, Senna~\cite{jiang2024senna}, OTTER~\cite{huang2025ottervisionlanguageactionmodeltextaware}, and AutoVLA~\cite{zhou2025autovla} exemplify this VLA paradigm, unifying perception, reasoning, and control in single autoregressive sequence models~\cite{sapkota2025vla, ma2024surveyvla}. These models are typically trained on large-scale, instruction-based datasets~\cite{fu2025orion, arai2025covla} to follow natural language commands and explain their decisions.

Despite these advances, current E2E systems still face significant challenges in long-horizon reasoning, safety guarantees, and interpretability. Recent works have begun to address these issues by integrating differentiable motion planning layers~\cite{chen2024ppad, weng2024paradrive} and constrained policy optimization techniques. Approaches like Model Predictive Control (MPC) integration~\cite{miyaoka2024chatmpc, long2024vlmmpc} and causal reasoning frameworks are also being explored. However, the performance of VLA models scales directly with the volume and diversity of real-world data\cite{naumann2025data, zheng2024preliminary}, highlighting the critical need for large-scale, multi-modal corpora like 4DLidarOpen.

Benchmarking in E2E driving has evolved alongside these advances, with simulators like CARLA~\cite{dosovitskiy2017carla} and large-scale real-world datasets such as nuScenes~\cite{nuscenes2019}, Waymo Open~\cite{waymo2024open}, and Argoverse~\cite{chang2019argoverse, wilson2023argoverse2} providing standardized evaluation platforms. Nevertheless, real-world deployment challenges remain significant due to domain gaps and the difficulty of certifying data-driven systems for safety-critical applications.

4DLidarOpen is specifically designed to support these emerging paradigms by providing precisely synchronized multi-modal sensor streams. The dataset facilitates research on unified perception-planning-control networks, advanced sensor fusion strategies, and predictive models that leverage motion information. By capturing real-world driving scenarios with rich annotations, 4DLidarOpen aims to bridge the gap between simulation-based research and practical deployment, accelerating progress toward more robust, safe, and interpretable E2E driving systems.

\subsection{4D FMCW Lidar}
Frequency-Modulated Continuous-Wave (FMCW) Lidar technology has recently advanced to deliver 4D point clouds $(x,y,z,v)$ at video frame rates, where $v$ represents the instantaneous radial velocity of each point~\cite{thornton2021fmcw}. This velocity channel provides a direct measurement of radial motion, distinguishing 4D FMCW Lidar from conventional time-of-flight Lidars that infer motion mainly through temporal association. However, the absence of large-scale, publicly available 4D datasets has limited high-level perception research using FMCW Lidar.

Existing FMCW Lidar research has predominantly focused on low-level tasks such as odometry and localization, leaving high-level perception relatively underexplored. HeLiPR~\cite{jung2024helipr} represents the first dedicated 4D FMCW Lidar dataset, designed specifically for inter-Lidar place recognition by filtering dynamic objects and performing place recognition using static points. The Doppler iterative closest point (DICP) algorithm~\cite{chen2023dicp} addresses the specific challenge of registering 4D FMCW Lidar point clouds by incorporating velocity consistency into the optimization process, improving accuracy in dynamic environments~\cite{yoneda2023extended}. Other works have explored odometry tasks, leveraging FMCW Lidar's ability to separate dynamic objects from static scenes to achieve more robust performance~\cite{wu2022picking, yoon2023need, zhao2024fmcw}. Recent advancements also include tightly coupled sensor fusion for odometry~\cite{zhao2024free} and efficient continuous-time trajectory estimation methods~\cite{9636676}.

This focus on low-level tasks leaves an important gap in high-level perception and driving benchmarks: there are few large-scale 4D FMCW datasets with high-level annotations for perception and driving tasks. Initial investigations into perception applications, such as the work by Gu et al.~\cite{gu2022iros}, demonstrate the potential of 4D FMCW Lidar for direct object tracking, benefiting from its instantaneous velocity measurements. Recent studies have begun to explore more complex perception tasks like 4D object detection~\cite{li2024perception}, highlighting the technology's advantages in motion estimation and velocity measurement for autonomous driving applications.

4DLidarOpen addresses this gap by providing the first large-scale corpus that couples 4D FMCW Lidar data with high-level perception and E2E driving annotations. By making this resource available to the research community, we aim to accelerate the development of motion-aware perception systems that can fully leverage the unique capabilities of 4D FMCW Lidar technology.
\section{Methodology}

\subsection{Overview}
4DLidarOpen is designed as a large-scale multi-modal benchmark for 4D scene understanding, 3D object detection, and E2E driving in complex urban environments. Compared with existing autonomous driving datasets, it emphasizes heterogeneous Lidar sensing and direct radial velocity measurements from 4D FMCW Lidar, providing a testbed for evaluating motion-aware perception and planning systems.

\subsection{Sensor Configuration}
\label{sec:sensors}

\begin{figure}[htbp]
    \centering
    \includegraphics[width=0.45\textwidth]{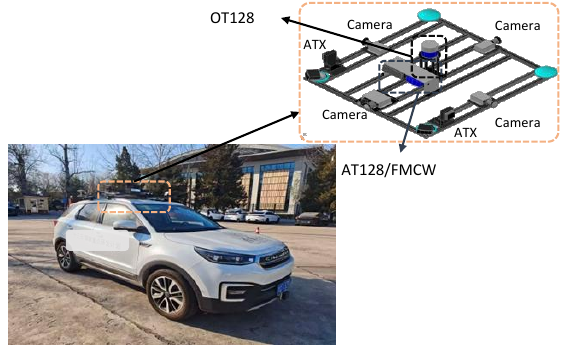}
    \caption{4DLidarOpen sensor configuration: five Lidars and five surround-view cameras mounted on the ego vehicle, providing comprehensive coverage around the vehicle.}
    \label{fig:sensor}
\end{figure}

4DLidarOpen employs a multi-modal sensor suite comprising five Lidars and five surround-view cameras, as illustrated in Figure \ref{fig:sensor}. The system captures 10-Hz Lidar sweeps synchronized with 20-Hz imagery from five cameras: four wide-angle cameras covering the sides and rear, and one telephoto front unit, together providing a panoramic field of view around the ego vehicle. For each sensor, 4DLidarOpen provides detailed camera intrinsics, extrinsics, and 6-DOF ego-vehicle poses in a global coordinate frame.

The sensor suite includes five distinct Lidars: a 4D FMCW Lidar, a 360° rotating OT Lidar, a solid-state AT Lidar, and two ATX blind-spot Lidars, all operating at 10 Hz. Global-shutter cameras are hardware-triggered to expose precisely while the Lidar sweeps across their field of view, ensuring sub-millisecond temporal alignment between sensor modalities.

\textbf{Synchronization.} 4DLidarOpen achieves camera–Lidar temporal alignment within $[-1.39,1.39]$ milliseconds, three times tighter than the $[-6,7]$ millisecond alignment of Waymo Open\cite{sun2020scalability}. This precise synchronization enables seamless pixel-point fusion without motion blur, critical for accurate multi-modal perception.

\subsection{Data Collection and Scenes}
\label{sec:collection}

\begin{figure*}[htbp]
    \centering
    \includegraphics[width=0.8\textwidth]{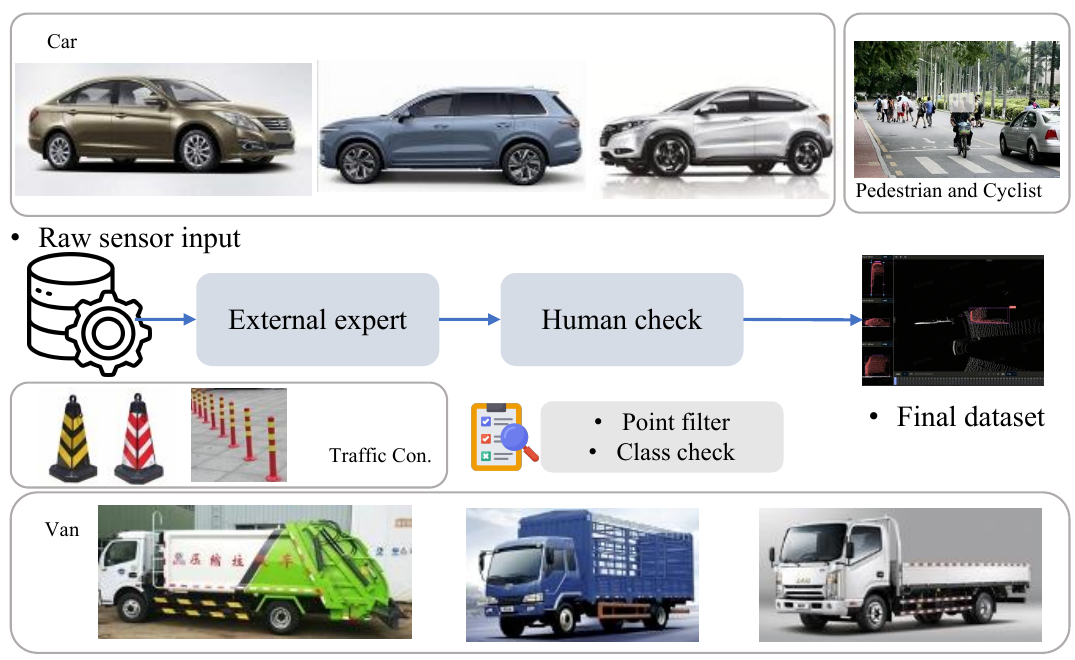}
    \caption{4DLidarOpen data processing pipeline, including raw data collection, sensor synchronization, automatic labeling, human verification, and final dataset generation.}
    \label{fig:data-pipe}
\end{figure*}

4DLidarOpen was recorded in Beijing's Yizhuang and Shougang Industrial Park districts, areas that encapsulate the complexity of contemporary Chinese urban traffic—from multi-lane arterials and highway interchanges to dense intersections and urban thoroughfares. Figure~\ref{fig:data-pipe} depicts the comprehensive processing pipeline, from initial data collection through synchronization, annotation, and validation.

The data collection campaign systematically sampled diverse driving scenarios to ensure comprehensive coverage:
\begin{itemize}
    \item \textbf{Road types:} multi-lane arterials, highway interchanges, high-speed segments, and urban streets with varying speed limits and traffic patterns.
    \item \textbf{Illumination conditions:} sunny, cloudy, dusk, dawn, and night scenarios to stress-test system robustness across different lighting environments.
\end{itemize}

The dataset is further enriched by frequent critical events that challenge autonomous driving systems:
\begin{itemize}
    \item \textbf{Pedestrian interactions:} dense crowds, jaywalking, and pedestrians crossing at non-designated locations.
    \item \textbf{High-speed cruising:} stable highway navigation at elevated speeds, testing long-range perception and prediction.
    \item \textbf{Unprotected maneuvers:} left turns and U-turns amid flowing traffic, requiring precise interaction prediction.
    \item \textbf{Congested traffic:} stop-and-go queues that stress close-range perception and low-speed control.
\end{itemize}

This diversity provides a challenging evaluation setting for perception, prediction, and planning under realistic urban traffic conditions.

\subsection{Annotations and Sample Format}
\label{sec:annotations}

4DLidarOpen provides rich, high-frequency 3D annotations to support a wide range of research tasks. Figure~\ref{fig:fmcw-sample} shows a representative sample with four camera views, Lidar point cloud, and human-annotated 3D boxes with persistent track IDs.


Every object within 4DLidarOpen's 5-class taxonomy is annotated with a precise 3D cuboid at 10 Hz, ensuring temporal consistency through unique track identifiers that persist across frames~\cite{Caesar20cvpr_NuScenes}. Our annotation policy is designed to maximize both relevance and precision: we annotate all objects within 150 meters of the ego vehicle, with particular focus on actionable obstacles that could impact driving decisions. Objects in the ego lane are tagged even with sparse Lidar evidence—sometimes as few as 5 points—ensuring that critical obstacles are not missed~\cite{Geiger12cvpr_KITTIAreWeReady,Behley19iccv_SemanticKITTI}. This curated approach, executed through a rigorous annotate-then-verify workflow on our internal platform, improves annotation reliability and ensures that the retained objects are relevant to driving decisions.

\textbf{Privacy.} To ensure compliance with privacy regulations, all faces and license plates in the dataset are automatically blurred during processing.

\textbf{Splits.} 4DLidarOpen provides 225 human-annotated scenarios, including 167 training scenarios and 58 validation scenarios, together with three auto-labeled training tiers containing 500, 1,000, and 2,000 sequences. This tiered structure enables scalable self-supervised research across different computational budgets.

To support large-scale self-supervised and semi-supervised learning, we also release a substantial collection of \textbf{auto-labeled} sequences across three distinct tiers: a \textbf{Small} set (500 scenarios), a \textbf{Medium} set (1,000 scenarios), and a \textbf{Large} set (2,000 scenarios). This flexible structure allows researchers to conduct experiments under varying computational constraints and data scales, facilitating everything from quick prototyping to large-scale pretraining.

\subsection{Data Processing and Labelling Policy}
\label{sec:processing}

4DLidarOpen supports a wide range of research tasks through a rigorously validated processing pipeline, enabling comprehensive evaluation of autonomous driving systems.

\textbf{Processing.} Raw PCAP Lidar streams are decoded to compact PLY format for efficient storage and processing. Ego-vehicle poses are fused from wheel odometry and IMU data to provide accurate positioning information. All sensors are synchronized using gPTP (general Precision Time Protocol) to achieve sub-millisecond accuracy. Data are rectified to a rear-axle-centered coordinate frame using pre-calibrated extrinsics, ensuring consistent spatial referencing across all sensor modalities. Sequences with anomalous poses or sensor artifacts are quarantined before annotation to maintain data quality.

\textbf{Annotation.} Our annotation pipeline begins with an auto-labeling system that generates preliminary 3D bounding boxes. These initial annotations are then iteratively refined by human experts through a two-stage review process. Objects within 150 meters of the ego vehicle that reflect more than 5 Lidar returns are systematically annotated, while ego-lane obstacles are retained even with sparser returns to ensure critical objects are captured. The two-stage review pipeline ensures every cuboid and track meets strict pixel-point fidelity standards, resulting in high-quality annotations for both training and evaluation.

\begin{figure*}[htbp]
    \centering
    \includegraphics[width=0.8\textwidth]{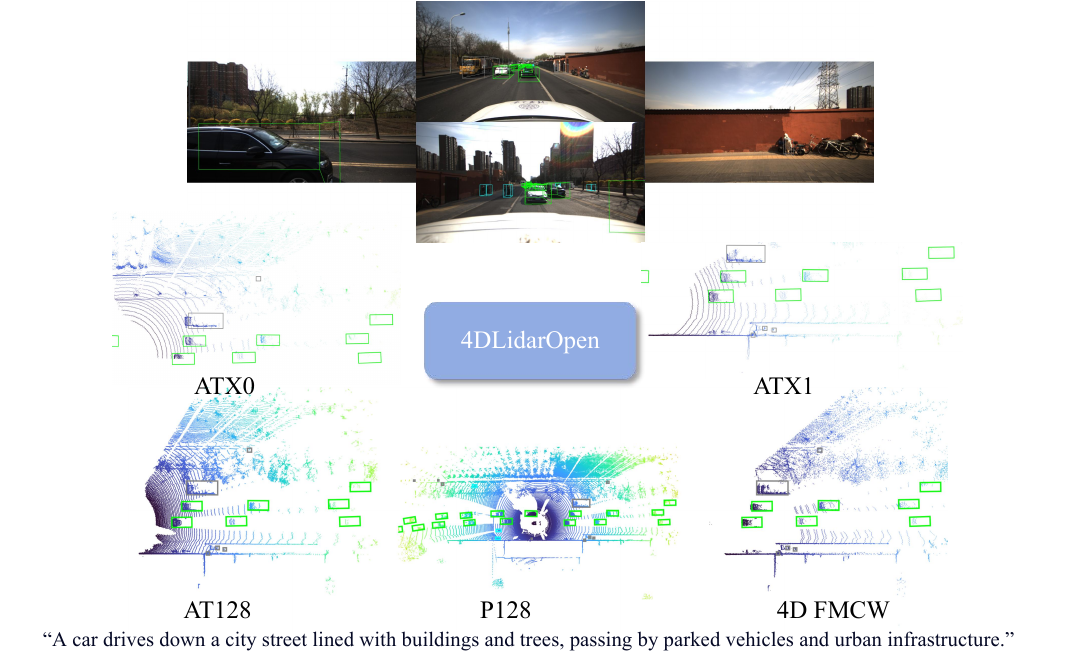}
    \caption{4DLidarOpen sample showing 4D FMCW Lidar data with velocity information: (a) raw point cloud with radial velocity coloring, (b) semantic segmentation results, (c) motion flow visualization with vector arrows, (d) comparison with conventional 3D Lidar without velocity cues. The 4D FMCW data provides point-wise radial velocity cues that support motion analysis and dynamic-object detection.}
    \label{fig:fmcw-sample}
\end{figure*}

Figure~\ref{fig:fmcw-sample} illustrates the advantages of 4D FMCW Lidar data in 4DLidarOpen. The instantaneous radial velocity information enables precise motion analysis, allowing for earlier detection and tracking of dynamic objects compared to conventional 3D Lidar systems.

\subsection{Data Statistics}
\label{sec:statistics}

Table~\ref{tab:dataset_stats} presents a comprehensive statistical overview of 4DLidarOpen's subsets, highlighting the balance between high-quality human annotations and large-scale auto-labeled data.

\begin{table*}[t]
\centering
\caption{STATISTICAL OVERVIEW OF THE THREE SUBSETS IN 4DLidarOPEN.}
\label{tab:dataset_stats}
\begin{tabular}{l|ccc}
\hline
\textbf{Statistic} & \textbf{Training (Human-Annot.)} & \textbf{Validation (Human-Annot.)} & \textbf{Training (Auto-Labeled)} \\ 
\hline
Number of Scenarios & 167 & 58 & 2,000 \\
Total Driving Distance (km) & 11.60 & 3.49 & 295.70 \\
Total Driving Duration (hours) & 0.58 & 0.19 & 10.89 \\
Total Camera Frames & 105,220 & 35,000 & 1,974,480 \\
Total Lidar Frames & 21,044 & 7,000 & 394,896 \\
Total Annotation Instances & 1,457,552 & 439,675 & 22,952,926 \\
Total Tracks & 730 & 592 & 1,741 \\
Avg. Annos. per Frame & 69.26 & 62.81 & 58.12 \\
\hline
\end{tabular}
\end{table*}

The human-annotated splits (225 scenes) yield an average of 69.3 and 62.8 instances per Lidar frame for training and validation respectively, with 730 and 592 distinct tracks. This rich temporal information enables robust modeling of object dynamics and interactions over time.

The auto-labeled tier (2,000 scenes, 295.7 km, 11 hours) delivers 1.97 million camera frames and 394,000 Lidar sweeps with 22.9 million bounding boxes, providing a massive dataset for scalable self-supervised learning and pretraining.

\begin{figure*}[htbp]
    \centering
    \includegraphics[width=0.9\textwidth]{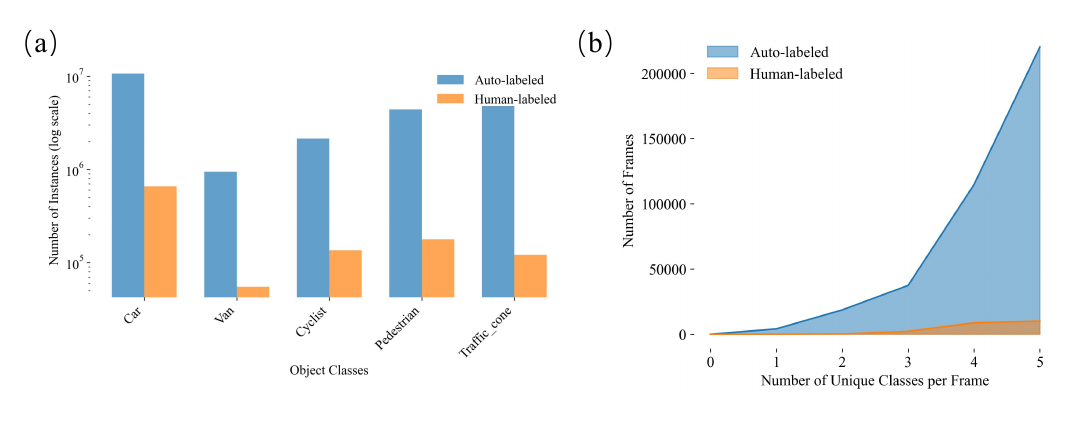}
    \caption{4DLidarOpen class and richness statistics. (a) Instance counts across five categories (Car, Van, Cyclist, Pedestrian, Traffic Cone) comparing auto-labeled (blue) and human-labeled (orange) annotations on log scale. (b) Distribution of unique class counts per Lidar frame, showing rich scene diversity.}
    \label{fig:stat-class}
\end{figure*}

Category statistics in Figure~\ref{fig:stat-class} show the distribution of annotations across object categories, comparing auto-labeled and human-labeled data. The long-tail distribution dominated by cars is evident, with auto-labeled data providing significantly more instances for rare categories like cyclists and traffic cones, supporting research on rare-event detection.

\begin{figure*}[htbp]
    \centering
    \includegraphics[width=0.9\textwidth]{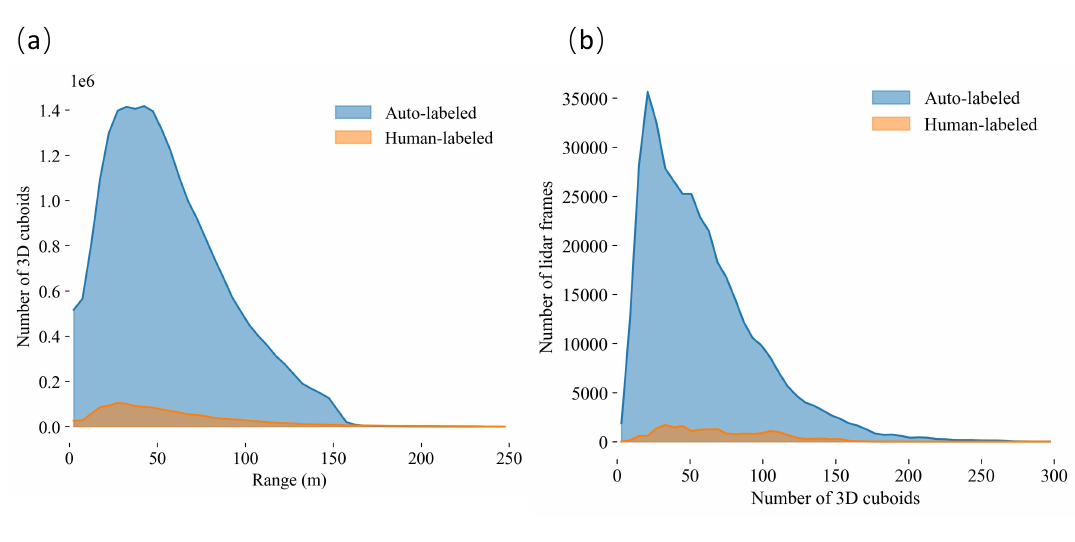}
    \caption{4DLidarOpen spatial and density statistics. (a) Object distance distribution showing peak concentration within 50 meters. (b) Distribution of 3D cuboid counts per Lidar frame, reflecting real-world traffic density variations.}
    \label{fig:stat-object}
\end{figure*}

Spatial analysis in Figure~\ref{fig:stat-object} shows that objects are concentrated within 50 meters of the ego vehicle, which aligns with typical Lidar effective range and highlights the importance of near-field perception for safe driving. The density distribution per Lidar frame reflects realistic urban traffic conditions.

\begin{figure*}[htbp]
    \centering
    \includegraphics[width=0.9\textwidth]{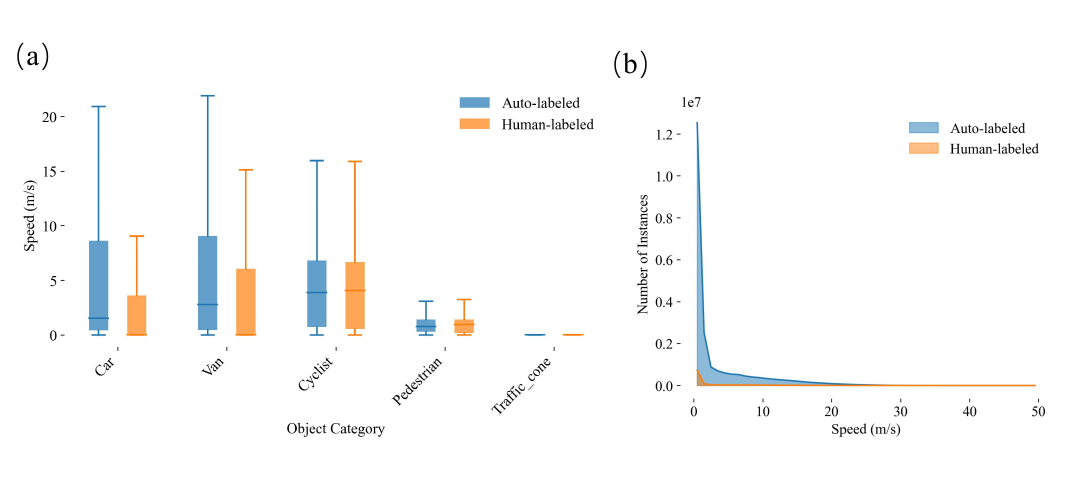}
    \caption{4DLidarOpen speed statistics. (a) Category-wise box plots showing speed distributions for Car, Van, Cyclist, Pedestrian, and Traffic Cone. (b) Overall speed histogram across all object instances.}
    \label{fig:stat-speed}
\end{figure*}

Speed profiles in Figure~\ref{fig:stat-speed} reveal distinct category-specific kinematic patterns: cars and vans exhibit higher speeds, while pedestrians and cyclists show lower velocity distributions. These statistics provide valuable insights for designing motion-aware perception models and velocity prediction systems.

Together, 4DLidarOpen's human-curated and auto-labeled tiers constitute a versatile and comprehensive resource for advancing research in 4D perception, motion prediction, and E2E driving systems. The dataset's scale, diversity, and rich annotations make it an ideal platform for developing and evaluating autonomous driving technologies.
\section{Experiments}

We evaluate 4DLidarOpen on three downstream tasks: 3D object detection, BEV segmentation and flow prediction, and motion forecasting with planning. 

We benchmark each task below.

\subsection{3D Object Detection}

\subsubsection{Metrics}
We adopt KITTI and nuScenes metrics for compatibility.

\textbf{KITTI AP:} IoU=0.5 for vehicles/vans; IoU=0.25 for vulnerable road users.
\begin{equation}
    \mathrm{AP}_\text{KITTI} = \frac{1}{41} \sum_{r \in \{0,0.025,\dots,1\}} p(r)
\end{equation}
where $p(r)$ is the precision interpolated at each recall level $r$ following the KITTI 40-point sampling scheme.

\textbf{nuScenes mAP/NDS:} distance-based mAP and NDS (aggregating translation, scale, orientation, velocity, and attribute errors).
mAP is computed over the distance-based matching strategy:
\begin{equation}
    \mathrm{mAP} = \frac{1}{C}\sum_{c=1}^{C} \frac{1}{D}\sum_{d=1}^{D} \mathrm{AP}_{c,d}
\end{equation}

NDS aggregates mAP with five true-positive error rates:
\begin{equation}
    \mathrm{NDS} = \frac{1}{10}\Bigl[5\times\mathrm{mAP} + \sum_{k\in\{\mathrm{trans},\mathrm{scale},\mathrm{orient},\mathrm{vel},\mathrm{attr}\}}(1-e_k)\Bigr]
\end{equation}
where $e_k$ denotes the average error for translation, scale, orientation, velocity, and attribute classification, each clipped to [0,1].

Five categories are evaluated: Car, Pedestrian, Cyclist, Van, and Traffic Cone. Vans are split by wheel-base and height; cones require $\geq 50\%$ above-ground visibility. All metrics are implemented in our open-source evaluation toolkit.

\subsubsection{Experimental Results}

Table~\ref{tab:baseline_det} reports the 3D object detection performance of different Lidar configurations using the SparseConv encoder and Sparse4D head. Table~\ref{tab:baseline_det_wt} further evaluates the contribution of the radial velocity channel for 4D FMCW Lidar-based detection.

We benchmark CenterPoint~\cite{Yin21cvpr_CenterPoint} and PointPillars~\cite{Lang19cvpr_PointPillars} backbones with a Sparse4D head, matching SparseDrive and DiffusionDrive architectures.

All Lidars use consistent voxelization parameters following KITTI/nuScenes conventions\cite{Lang19cvpr_PointPillars,Yin21cvpr_CenterPoint}, with 4D FMCW providing additional instantaneous radial velocity $v_r$\cite{jung2024helipr}.

\paragraph{Analysis:} 
Table~\ref{tab:baseline_det} presents the 3D object detection results across three Lidar sensors, revealing distinct performance characteristics tied to sensor architecture.

\textbf{Overall Performance.} AT128 achieves the highest overall performance with mAP of 69.62\% and NDS of 76.32\%, establishing the strongest baseline for detection tasks. OT128 follows closely in NDS (74.51) but lags in mAP (64.89), while 4D FMCW exhibits the lowest raw detection metrics (mAP: 63.62, NDS: 72.05) despite its velocity-sensing capabilities.

\textbf{Detailed Category Analysis.} 
AT128 demonstrates superior performance in detecting small and geometrically complex objects, particularly excelling in Traffic Cone (66.48\%) and Cyclist (86.87\%) detection, which suggests its scanning pattern provides denser point coverage for low-profile objects. 
OT128 achieves the highest Car detection accuracy (91.21\%), likely benefiting from its optimized beam distribution for mid-range vehicle detection.
4D FMCW shows competitive performance on Vans (57.08\%) and Pedestrians (76.81\%), but Traffic Cone detection remains more challenging for FMCW 4D (35.81\%), indicating that the velocity channel's spatial resolution trade-off particularly affects small static object detection.
\begin{table*}
    \centering
    \caption{3D object detection performance comparison across different Lidar sensors using SparseConv encoder and Sparse4D head. T.C. refers to Traffic Cone.}
    \resizebox{1.0\textwidth}{!}{
        \begin{tabular}{l|ll|ccccc|cc}
            \toprule
            Sensor & Encoder & Decoder  & $Car$ & $Ped.$ & $Cyc.$ & $Van$ & $T.C.$ & mAP & NDS \\
            \midrule
            AT128 & SparseConv   & Sparse4D head & 90.45 & 76.21 & 86.87 & 51.49 & 66.48 & 69.62 & 76.32 \\            
            FMCW 4D & SparseConv   & Sparse4D head & 87.87 & 76.81 & 71.63 & 57.08 & 35.81 & 63.62 & 72.05  \\            
            OT128 & SparseConv   & Sparse4D head & 91.21 & 70.51 & 81.90 & 54.79 & 53.80 & 64.89 & 74.51 \\            
            \bottomrule
        \end{tabular}
    }
    \label{tab:baseline_det}
\end{table*}

\begin{table*}
    \centering
    \caption{Ablation study on the impact of radial velocity ($v_r$) channel for 4D FMCW Lidar-based 3D detection. T.C. refers to Traffic Cone.}
    \resizebox{1.0\textwidth}{!}{
        \begin{tabular}{l|ll|ccccc|cc}
            \toprule
            Sensor & Encoder & Decoder  & $Car$ & $Ped.$ & $Cyc.$ & $Van$ & $T.C.$ & mAP & NDS \\
            \midrule
            FMCW 4D xyzi & SparseConv   & Sparse4D head & 22.57 & 17.59 & 40.98 & 19.59 & - & 21.58 & 41.86 \\            
            FMCW 4D xyziv & SparseConv   & Sparse4D head & 34.03 & 48.28 & 60.45 & 31.05 & - & 28.75 & 51.49  \\    
            \bottomrule
        \end{tabular}
    }
    \label{tab:baseline_det_wt}
\end{table*}

\paragraph{Velocity Channel Ablation:} 
Table~\ref{tab:baseline_det_wt} shows the contribution of the radial velocity channel to detection performance. 
In the absence of $v_r$ (xyzi only), the detection performance decreases across all categories, with mAP and NDS dropping to 21.58\% and 41.86\%, respectively. 
Incorporating velocity information (xyziv) improves mAP from 21.58\% to 28.75\% and NDS from 41.86\% to 51.49\%. 
The improvement is particularly evident for pedestrians, where AP increases from 17.59\% to 48.28\%, suggesting that motion cues are useful for detecting dynamic vulnerable road users. 
These results confirm that while 4D FMCW sacrifices some spatial resolution compared to scanning Lidars, the instantaneous velocity channel provides complementary information that improves detection performance in dynamic scenes.

\subsection{Bird's Eye View Segmentation and Flow}
We benchmark BEV segmentation and flow on 4DLidarOpen against MotionNet\cite{wu2020motionnet}, BE-STI\cite{wang2022sti}, PriorMotion\cite{qian2025priormotion}, and LEGO-Motion\cite{qian2025lego}. Given Lidar sequence $\{\mathcal{P}_t\}_{t=1}^{T}$, we learn $f$ that forecasts BEV motion $\mathcal{M}_t$, classifies cells $\mathcal{C}_t$, and estimates static probability $\mathcal{S}_t$.
\begin{equation}
    f(\{\mathcal{P}_t\}_{t=1}^{T}) \rightarrow (\mathcal{M}_t, \mathcal{C}_t, \mathcal{S}_t)
\end{equation}

Input Lidar clouds $\mathcal{P}_t\!=\!\{P_t^i\}_{i=1}^{N_t}$ are voxelized to $\mathcal{V}_t\!\in\!\{0,1\}^{H\times W\times C}$ in ego coordinates.

The model outputs BEV motion field $\mathcal{M}_t$, class logits $\mathcal{C}_t$, and static probability $\mathcal{S}_t$ per cell:
\vspace{-0.5em}
\begin{equation}
\mathcal{M}_t\in\mathbb{R}^{H\times W\times2},\; \mathcal{C}_t\in\mathbb{R}^{H\times W\times N_c},\; \mathcal{S}_t\in\mathbb{R}^{H\times W}.
\end{equation}

   \begin{figure*}[htbp]
    \centering
    \includegraphics[width=0.9\textwidth]{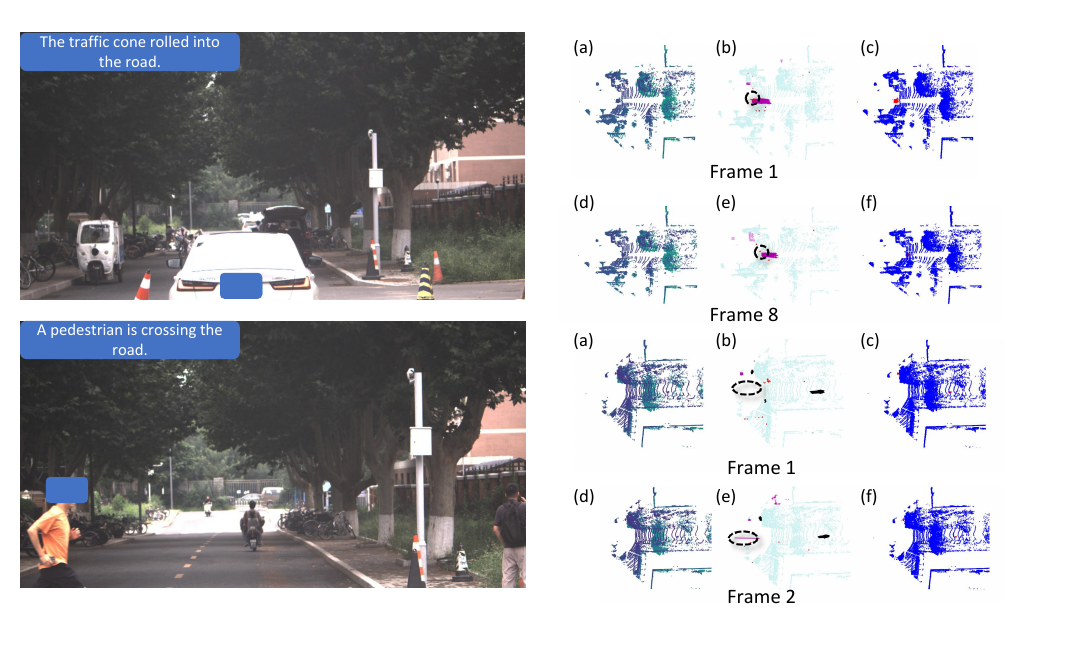}
    \caption{4DLidarOpen campus ablation experiment. Top row: rolling cone scenario; bottom row: darting pedestrian scenario. (a)-(c) 4D FMCW Lidar results showing raw BEV point cloud, semantic grid, and radial velocity heatmap. (d)-(f) Baseline results without velocity information. 4D FMCW detects vulnerable road users (VRUs) significantly earlier (cone: frame 1 vs 8; pedestrian: frame 1 vs 2), illustrating the value of instantaneous velocity cues.}
    \label{fig:visual-campus}
\end{figure*}

\paragraph{Campus Ablation Experiment Analysis.}
Figure~\ref{fig:visual-campus} illustrates the potential benefit of 4D FMCW Lidar for early dynamic-object detection. In the rolling cone scenario (top row), the 4D FMCW Lidar detects the falling cone in frame 1, while the baseline without velocity information requires 8 frames to achieve the same detection. Similarly, in the darting pedestrian scenario (bottom row), the 4D FMCW Lidar identifies the pedestrian in frame 1, compared to frame 2 for the baseline. This early detection capability is critical for autonomous driving systems to react in time to sudden events. The radial velocity heatmap (c) clearly shows the motion of the cone and pedestrian, providing qualitative evidence of the velocity channel's contribution to early detection.

\begin{figure*}[htbp]
    \centering
    \includegraphics[width=0.9\textwidth]{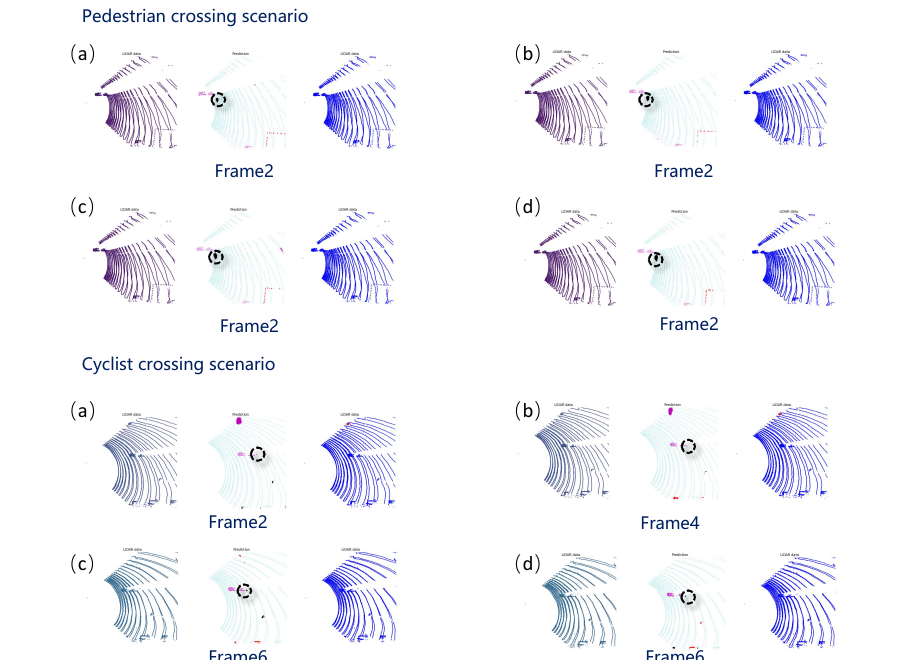}
    \caption{4DLidarOpen Tianjin crossing test. Top row: pedestrian crossing scenario; bottom row: e-bike crossing scenario. (a) 4D FMCW Lidar + our model; (b) 3D Lidar + our model; (c) 4D FMCW Lidar + baseline model; (d) 3D Lidar + baseline model. 4D input significantly advances detection (e-bike: frame 2 vs 6) and stabilizes earlier (frame 4 vs 29), showing the benefit of direct velocity measurements in crossing scenarios.}
    \label{fig:tianjin-crossing}
\end{figure*}

\paragraph{Tianjin Crossing Test Analysis.}
Figure~\ref{fig:tianjin-crossing} further demonstrates the performance of 4D FMCW Lidar in challenging crossing scenarios. In both pedestrian and e-bike crossing scenarios, the 4D FMCW Lidar configurations (a and c) consistently outperform their 3D Lidar counterparts (b and d). Specifically, in the e-bike crossing scenario, the 4D FMCW Lidar + our model (a) detects the e-bike in frame 2, while the 3D Lidar + our model (b) requires frame 6. Moreover, the 4D configuration stabilizes its detection by frame 4, compared to frame 29 for the 3D configuration. This improvement in detection speed and stability suggests that velocity information can support timely decision-making in dynamic crossing scenarios.

\subsubsection{Metrics}

Following established evaluation protocols, we group cells by speed: static ($\leq$0.2 m/s), slow (0.2--5 m/s), and fast ($>$5 m/s), and report mean and median L2 displacement errors at 1 second. For each speed group, we calculate the average distance between predicted and ground truth displacements. The mean prediction error for a group $G$ is:
\begin{equation}
    \text{Mean Error}_G = \frac{1}{|G|} \sum_{i \in G} \| \mathbf{\hat{d}}_i - \mathbf{d}_i \|_2
\end{equation}
where \( \mathbf{\hat{d}}_i \) is the predicted displacement and \( \mathbf{d}_i \) is the ground truth displacement for cell \( i \).

The median prediction error for a group \( G \) is:
\begin{equation}
    \text{Median Error}_G = \text{median} \left( \left\{ \| \mathbf{\hat{d}}_i - \mathbf{d}_i \|_2 \mid i \in G \right\} \right)
\end{equation}

To better understand motion prediction performance, we decompose the displacement error along radial and lateral directions relative to the ego vehicle. The radial direction corresponds to motion toward or away from the ego vehicle, while the lateral direction represents side-to-side motion.

We classify cells into static, slow, and fast groups based on their radial and lateral speeds, respectively, and compute the mean and median errors in each direction. The mean radial prediction error for group \( G_r \) is:
\begin{equation}
    \text{Mean Radial Error}_{G_r} = \frac{1}{|G_r|} \sum_{i \in G_r} \left| (\mathbf{\hat{d}}_i - \mathbf{d}_i) \cdot \mathbf{u}_r \right|
\end{equation}
where \( \mathbf{u}_r \) is the unit vector pointing from the ego vehicle to the cell center.

Similarly, the mean lateral prediction error for group \( G_l \) is:
\begin{equation}
    \text{Mean Lateral Error}_{G_l} = \frac{1}{|G_l|} \sum_{i \in G_l} \left| (\mathbf{\hat{d}}_i - \mathbf{d}_i) \cdot \mathbf{u}_l \right|
\end{equation}
where \( \mathbf{u}_l \) is the unit vector orthogonal to \( \mathbf{u}_r \) in the horizontal plane.

The corresponding median errors are computed analogously for both radial and lateral directions.

In addition to motion prediction, we evaluate cell classification performance using overall accuracy (OA) and mean category accuracy (MCA). OA measures the average accuracy across all non-empty cells, while MCA calculates the average accuracy per category, providing insight into how well models handle different object types.


\subsubsection{Experimental Analysis}
\label{sec:exp_analysis}

Table~\ref{tab:speed_classification} summarizes the BEV segmentation and flow estimation results across three architectures (MotionNet, STI, PriorMotion) and three sensor configurations (OT128, AT128, FMCW 4D), with and without velocity input fusion.

\textbf{Overall Performance.} 
PriorMotion combined with FMCW 4D and velocity fusion achieves the best overall performance among the evaluated configurations, with the lowest motion prediction errors across all speed regimes (static: 0.0011, slow: 0.6218, fast: 1.0033) and the highest classification accuracy (MCA: 0.8541, OA: 0.9864). 
The consistent superiority of velocity-enhanced variants (denoted by \textdagger) across all model-sensor combinations indicates the importance of explicit motion cues for flow estimation. 
Notably, the incorporation of velocity information reduces static cell errors by an order of magnitude (from $\sim$0.01 to $\sim$0.001), indicating precise ego-motion compensation and static background stabilization.

\textbf{Architecture Comparison.} 
PriorMotion consistently outperforms MotionNet and STI across all sensor configurations, with the performance gap widening for fast-moving objects (PriorMotion+FMCW 4D\_0+v achieves 1.0033 vs. MotionNet's 1.0844 and STI's 1.0391). 
This suggests that PriorMotion's motion-aware architectural priors better capture dynamic scene evolution compared to convolutional (MotionNet) or spatial-temporal interaction (STI) approaches. 
STI demonstrates competitive lateral motion estimation (best lateral slow error: 0.3916 with STI+FMCW 4D\_0+v), likely benefiting from its explicit spatial-temporal feature interaction mechanism.

\textbf{Sensor Analysis.} 
FMCW 4D exhibits distinct performance characteristics depending on velocity fusion: without velocity input, it generally underperforms compared to scanning Lidars (OT128/AT128), particularly in lateral motion estimation (FMCW 4D\_0 lateral slow error: 0.4632 vs. AT128\_0: 0.4473 for MotionNet). 
However, with velocity fusion, FMCW 4D configurations achieve the best radial motion estimation (best radial slow: 0.4478, best radial fast: 0.6385) and competitive lateral performance, leveraging the instantaneous Doppler measurements for direct motion decomposition. 
AT128 shows strong raw performance in slow and fast motion regimes without velocity fusion (best non-v slow: 0.6385, best non-v fast: 0.9959), attributed to its dense point cloud providing rich geometric cues for optical flow-style motion estimation.

\textbf{Directional Motion Decomposition.} 
The decomposition of motion errors into radial and lateral components reveals that velocity fusion predominantly improves radial motion estimation (reducing radial static errors by 15-20$\times$), while lateral motion benefits are more modest but consistent. 
This asymmetry reflects the physical nature of FMCW Doppler measurements, which directly observe radial velocity but require geometric inference for lateral components. 
PriorMotion+FMCW 4D\_0+v achieves the best lateral fast error (0.7093), suggesting that advanced architectures can effectively disambiguate lateral motion from radial Doppler cues and spatial context.

\begin{table*}[htbp]
    \centering
    \caption{Comparison of BEV segmentation and flow performance across different models and sensors (The smaller the value, the lower the speed error. $\downarrow$: the lower the better for error metrics; $\uparrow$: the higher the better for classification metrics. Static: speed $\leq$ 0.2~m/s, Slow: 0.2~m/s $<$ speed $\leq$ 5~m/s, Fast: speed $>$ 5~m/s. OA: Overall Accuracy, MCA: Mean Category Accuracy. \textdagger~denotes the use of the radial velocity channel).}
    \resizebox{0.95\textwidth}{!}{%
        \begin{tabular}{l|l|ccc|ccc|ccc|cc}
            \toprule
            Experiment & Sensor & \multicolumn{3}{c|}{Overall Speed $\downarrow$} & \multicolumn{3}{c|}{Radial Speed $\downarrow$} & \multicolumn{3}{c|}{Lateral Speed $\downarrow$} & \multicolumn{2}{c}{Grid Classification $\uparrow$} \\
            & & \multicolumn{3}{c|}{Classification Interval} & \multicolumn{3}{c|}{Classification Interval} & \multicolumn{3}{c|}{Classification Interval} &   &  \\
            \midrule
            & & Static & Slow & Fast & Static & Slow & Fast & Static & Slow & Fast & OA &  MCA \\
            \midrule
            MotionNet\cite{wu2020motionnet}        & OT128\_1 & 0.0099 & 0.7103 & 1.1664 & 0.0057 & 0.5026 & 0.7779 & 0.0064 & 0.4424 & 0.8316 & 0.9825 & 0.7276 \\
            MotionNet        & AT128\_0 & 0.0098 & 0.7320 & 1.2625 & 0.0065 & 0.5100 & 0.7451 & 0.0060 & 0.4473 & 0.8290 & 0.9827 & 0.7364 \\
            MotionNet      & FMCW 4D\_0 & 0.0097 & 0.7328 & 1.2504 & 0.0063 & 0.4909 & 0.8021 & 0.0059 & 0.4632 & 0.8568 & 0.9817 & 0.7179 \\
            MotionNet\textsuperscript{\textdagger}   & FMCW 4D\_0 & \underline{0.0007} & 0.6805 & 1.0844 & \underline{0.0004} & 0.4656 & 0.6543 & \underline{0.0003} & 0.4204 & 0.7377 & 0.9817 & 0.7022 \\
            STI\cite{wang2022sti}             & OT128\_1 & 0.0097 & 0.6772 & 1.1482 & 0.0065 & 0.4732 & 0.6668 & 0.0060 & 0.4071 & 0.7751 & 0.9817 & 0.7341 \\
            STI            & AT128\_0 & 0.0099 & 0.6794 & 1.0464 & 0.0067 & 0.4533 & 0.6491 & 0.0061 & 0.4117 & 0.7049 & 0.9825 & 0.7375 \\
            STI           & FMCW 4D\_0 & 0.0099 & 0.7199 & 1.2354 & 0.0064 & 0.4891 & 0.7893 & 0.0060 & 0.4429 & 0.8600 & 0.9809 & 0.7128 \\
            STI\textsuperscript{\textdagger}         & FMCW 4D\_0 & \textbf{0.0006} & 0.6353 & 1.0391 & \textbf{0.0005} & \underline{0.4204} & \underline{0.6385} & \textbf{0.0004} & 0.3916 & \underline{0.7031} & 0.9821 & 0.7387 \\
            PriorMotion\cite{qian2025priormotion}     & OT128\_1 & 0.0098 & 0.6970 & 1.0517 & 0.0079 & 0.4836 & 0.6447 & 0.0073 & 0.4216 & 0.7174 & 0.9813 & 0.7345 \\
            PriorMotion      & AT128\_0 & 0.0099 & \underline{0.6385} & \underline{0.9959} & 0.0090 & 0.4545 & 0.6819 & 0.0044 & \underline{0.3873} & 0.7482 & 0.9804 & \underline{0.7572} \\
            PriorMotion    & FMCW 4D\_0 & 0.0103 & 0.6783 & 1.0801 & 0.0083 & 0.4886 & 0.6992 & 0.0077 & 0.4090 & 0.7308 & \underline{0.9867} & 0.8219 \\
            PriorMotion\textsuperscript{\textdagger}  & FMCW 4D\_0 & 0.0011 & \textbf{0.6218} & \textbf{1.0033} & 0.0009 & \textbf{0.4478} & 0.6829 & 0.0008 & \textbf{0.3822} & \textbf{0.7093} & \textbf{0.9864} & \textbf{0.8541} \\
            \bottomrule
        \end{tabular}
    }
    \label{tab:speed_classification}
\end{table*}

\textbf{Classification Performance.} 
While all configurations achieve high overall accuracy (OA $>$ 0.98), mean category accuracy (MCA) reveals significant disparities in handling rare or challenging semantic classes. 
PriorMotion+FMCW 4D\_0+v achieves MCA of 0.8541, substantially outperforming the next best configuration (PriorMotion+FMCW 4D\_0: 0.8219) and baseline scanning Lidar setups (AT128\_0: 0.7572). 
This 13\% relative improvement in per-class accuracy, coupled with superior motion estimation, indicates that velocity-informed models achieve more robust and semantically consistent scene understanding, critical for downstream prediction and planning tasks.

In summary, the experimental results indicate that \textbf{(1)} explicit velocity fusion is important for exploiting the motion-sensing capability of FMCW 4D Lidar, \textbf{(2)} PriorMotion's architectural design effectively uses motion priors across different sensor types, and \textbf{(3)} combining motion-aware architectures with 4D FMCW sensing improves both motion estimation and semantic classification in dynamic driving scenes.

\subsection{Motion Forecasting and Planning}
End-to-end trajectory prediction and planning are core components for driving models.
We adopt two state-of-the-art multi-modal predictors and planners as baseline E2E systems: (1) SparseDrive: a sparse query transformer that decodes agent-centric trajectories with cross-attention between agent and ego queries and feature maps. (2) DiffusionDrive: a diffusion-based policy that progressively denoises a set of waypoint latents conditioned on the driving command and surrounding agents. To fit for Lidar inputs, both methods are slightly adjusted to perform deformable cross attention on BEV feature maps.

\subsubsection{Metrics} 
The task is formulated as, given the current 1-second sensor context, each model forecasts the ego-vehicle and moving obstacles for the next 3 s (30 frames @ 10 Hz). Evaluation is restricted to dynamic objects (speed $>$ 1.0 m/s within the prediction horizon).

$minADE_k$: minimum Average Displacement Error over k predicted trajectories
\begin{equation}
    \operatorname{minADE}_k = \frac{1}{T} \sum_{t=1}^{T} \left\| \mathbf{p}^{*}_t - \hat{\mathbf{p}}^{(\text{best})}_t \right\|_2,
\quad T = 30
\end{equation}

$minFDE_k$: minimum Final Displacement Error at the last frame (t = 30)
\begin{equation}
    \operatorname{minFDE}_k = \left\| \mathbf{p}^{*}_{30} - \hat{\mathbf{p}}^{(\text{best})}_{30} \right\|_2
\end{equation}

We report results for k = 6 (full multi-modal output) and k = 1 (best-of-one trajectory) to quantify both diversity and single-hypothesis accuracy.

\subsubsection{Baseline Results} 

\begin{table*}
    \centering
    \caption{End-to-end motion forecasting results comparing different Lidar sensors with SparseDrive architecture.}
    \resizebox{1.0\textwidth}{!}{
        \begin{tabular}{l|l|cccc|cccc}
            \toprule
            Sensor & Method &  ADE(6) &  FDE(6) &  AHE(6) & FHE(6) & ADE(1) &  FDE(1) &  AHE(1) & FHE(1)\\
            \midrule
            AT128 & SparseDrive & 1.58 & 2.87 & 0.67 & 0.95 & 3.98 & 7.65 & 0.70 & 0.99 \\            
            FMCW 4D & SparseDrive & \textbf{1.47} & \textbf{2.58} & \textbf{0.63} & \textbf{0.89} & \textbf{3.58} & \textbf{6.85} & \textbf{0.65} & \textbf{0.93} \\            
            OT128 & SparseDrive & 1.98 & 3.61 & 0.72 & 0.99 & 5.20 & 9.91 & 0.77 & 1.05 \\
            \bottomrule
        \end{tabular}
    }
    \label{tab:baseline_predict}
\end{table*}

\begin{table*}
    \centering
    \caption{End-to-end planning results comparing different Lidar sensors with SparseDrive architecture.}
    \resizebox{1.0\textwidth}{!}{
        \begin{tabular}{l|l|cccc|cccc}
            \toprule
            Sensor & Method &  ADE(6) &  FDE(6) &  AHE(6) & FHE(6) & ADE(1) &  FDE(1) &  AHE(1) & FHE(1)\\
            \midrule
            AT128 & SparseDrive & 0.40 & 0.95 & 0.31 & 0.60 & 0.68 & 1.70 & 0.31 & 0.61 \\            
            FMCW 4D & SparseDrive & \textbf{0.38} & \textbf{0.79} & \textbf{0.27} & \textbf{0.52} & \textbf{0.66} & \textbf{1.54} & \textbf{0.25} & \textbf{0.51} \\            
            OT128 & SparseDrive & 0.57 & 1.01 & 0.30 & 0.64 & 0.95 & 1.92 & 0.25 & 0.53 \\
            \bottomrule
        \end{tabular}
    }
    \label{tab:baseline_plane}
\end{table*}

Table~\ref{tab:baseline_predict} and Table~\ref{tab:baseline_plane} present the E2E motion forecasting and planning results, revealing a different performance trend compared to the detection task.

\textbf{Overall Performance.} 
Contrary to the 3D detection results where AT128 achieved the highest mAP/NDS, \textbf{4D FMCW Lidar achieves the best performance among the evaluated Lidar configurations in both motion forecasting and planning tasks despite its lower raw detection metrics}. 
For motion forecasting (Table~\ref{tab:baseline_predict}), FMCW 4D achieves the lowest errors across all metrics: ADE(6) of 1.47m (7.0\% better than AT128, 25.8\% better than OT128) and FDE(6) of 2.58m (10.1\% better than AT128, 28.5\% better than OT128). 
The single-mode performance (k=1) shows even larger gains, with FMCW 4D achieving FDE(1) of 6.85m vs. AT128's 7.65m and OT128's 9.91m, indicating superior trajectory consistency even when the model commits to a single hypothesis.

For planning (Table~\ref{tab:baseline_plane}), FMCW 4D maintains its advantage with FDE(6) of 0.79m (16.8\% better than AT128, 21.8\% better than OT128) and FDE(1) of 1.54m (9.4\% better than AT128, 19.8\% better than OT128). 
Heading estimation errors (AHE/FHE) follow the same pattern, with FMCW 4D achieving the lowest angular deviations.

\textbf{Performance Gap Analysis.} 
The performance hierarchy consistently follows: FMCW 4D $>$ AT128 $>$ OT128 across all forecasting and planning metrics. 
The gap between FMCW 4D and scanning Lidars widens for long-horizon predictions (FDE vs. ADE) and single-mode evaluation (k=1 vs. k=6), suggesting that velocity information is particularly critical for maintaining trajectory consistency over time and reducing multi-modal ambiguity. 
OT128 exhibits notably degraded performance in forecasting (FDE(6): 3.61m vs. FMCW's 2.58m), likely due to sparser point density affecting motion estimation for distant dynamic agents.

\textbf{Implications for System Design.} 
These results suggest that higher object-detection accuracy does not necessarily translate into better downstream forecasting and planning performance. 
While AT128 excels at static geometric reconstruction (as evidenced by superior detection AP), FMCW 4D's instantaneous velocity measurements provide critical dynamic scene understanding that proves more valuable for motion forecasting and planning. 
The velocity channel enables:
\begin{itemize}
    \item \textbf{Early detection of dynamic agents:} Direct velocity observation reduces reliance on multi-frame motion estimation and can improve response latency in dynamic-object detection (illustrated in Figure~\ref{fig:visual-campus} and Figure~\ref{fig:tianjin-crossing}).
    \item \textbf{Robust motion initialization:} More informative initial velocity estimates can reduce the accumulation of errors in trajectory rollouts, explaining the superior long-horizon FDE performance.
    \item \textbf{Reduced multi-modality:} Velocity cues help reduce motion ambiguity and improve single-hypothesis prediction accuracy (k=1 metrics).
\end{itemize}

\subsection{Discussion}

The experimental results reveal a task-dependent relationship between geometric precision and direct motion measurement in Lidar sensing.

\textbf{Task-Specific Sensor Characteristics.} For 3D object detection, scanning Lidars such as AT128 show advantages in static geometric reconstruction and small-object detection, which can be attributed to denser spatial sampling. In contrast, 4D FMCW Lidar provides point-wise radial velocity measurements that are particularly beneficial for tasks involving dynamic-object motion, including BEV flow prediction, motion forecasting, and planning.

\textbf{Value of Direct Velocity Measurement.} The velocity-fused variants in Table~\ref{tab:speed_classification} consistently improve motion estimation, especially in the radial direction. This observation is consistent with the measurement principle of FMCW Lidar, which directly observes radial velocity while lateral motion still requires inference from spatial context. The ablation in Table~\ref{tab:baseline_det_wt} further shows that removing the velocity channel degrades detection performance, indicating that velocity provides complementary information rather than merely redundant features.

\textbf{Sensor and Architecture Co-Design.} The results also suggest that the benefit of 4D FMCW Lidar depends on the downstream model architecture. Motion-aware models such as PriorMotion better exploit velocity cues than generic perception backbones, indicating that future 4D Lidar systems should consider joint design of sensing modality, representation, and learning architecture.

\textbf{Practical Implications.} These findings suggest that 4D FMCW Lidar can complement scanning Lidars in autonomous driving systems. Scanning Lidars remain useful for dense geometric reconstruction, whereas 4D FMCW Lidar provides direct motion measurements for dynamic-scene understanding. Future multi-Lidar fusion strategies may benefit from explicitly combining these complementary sensing properties.





\section{Conclusion}

This paper presented 4DLidarOpen, an open large-scale multi-modal autonomous driving dataset centered on 4D FMCW Lidar sensing. The dataset integrates a forward-facing 4D FMCW Lidar, rotating OT Lidar, solid-state AT Lidar, ATX blind-spot Lidars, synchronized surround-view cameras, and 6-DOF ego-vehicle poses. It provides a benchmark platform for 4D scene understanding, multi-Lidar fusion, and motion-aware autonomous driving.

Experiments on 3D object detection, BEV segmentation and flow prediction, and motion forecasting with planning show that point-wise radial velocity measurements provide complementary information to conventional geometric Lidar sensing. While scanning Lidars remain advantageous for dense geometric reconstruction, 4D FMCW Lidar improves motion-related perception and downstream forecasting and planning, especially in dynamic scenarios.

The current dataset has several limitations. Its data are mainly collected in Beijing urban environments, and part of the training data relies on auto-labeled annotations that may contain residual tracking errors. Future work will extend the dataset to more cities, weather conditions, and sensor modalities, and will introduce long-horizon forecasting benchmarks and additional pretrained baselines.

4DLidarOpen provides a foundation for future research on direct motion measurement, velocity-aware perception, and robust planning for autonomous driving.



\bibliographystyle{IEEEtran}
\bibliography{reference.bib}

@inproceedings{jiang2025survey,
  title={A survey on vision-language-action models for autonomous driving},
  author={Jiang, Sicong and Huang, Zilin and Qian, Kangan and Luo, Ziang and Zhu, Tianze and Zhong, Yang and Tang, Yihong and Kong, Menglin and Wang, Yunlong and Jiao, Siwen and others},
  booktitle={Proceedings of the IEEE/CVF International Conference on Computer Vision},
  pages={4524--4536},
  year={2025}
}

@inproceedings{zheng2025world4drive,
  title={World4drive: End-to-end autonomous driving via intention-aware physical latent world model},
  author={Zheng, Yupeng and Yang, Pengxuan and Xing, Zebin and Zhang, Qichao and Zheng, Yuhang and Gao, Yinfeng and Li, Pengfei and Zhang, Teng and Xia, Zhongpu and Jia, Peng and others},
  booktitle={Proceedings of the IEEE/CVF International Conference on Computer Vision},
  pages={28632--28642},
  year={2025}
}

@inproceedings{min2024driveworld,
  title={Driveworld: 4d pre-trained scene understanding via world models for autonomous driving},
  author={Min, Chen and Zhao, Dawei and Xiao, Liang and Zhao, Jian and Xu, Xinli and Zhu, Zheng and Jin, Lei and Li, Jianshu and Guo, Yulan and Xing, Junliang and others},
  booktitle={Proceedings of the IEEE/CVF conference on computer vision and pattern recognition},
  pages={15522--15533},
  year={2024}
}

@inproceedings{shi2024streamingflow,
  title={StreamingFlow: Streaming occupancy forecasting with asynchronous multi-modal data streams via neural ordinary differential equation},
  author={Shi, Yining and Jiang, Kun and Wang, Ke and Li, Jiusi and Wang, Yunlong and Yang, Mengmeng and Yang, Diange},
  booktitle={Proceedings of the IEEE/CVF Conference on Computer Vision and Pattern Recognition},
  pages={14833--14842},
  year={2024}
}

@INPROCEEDINGS{9636676,
  author={Lv, Jiajun and Hu, Kewei and Xu, Jinhong and Liu, Yong and Ma, Xiushui and Zuo, Xingxing},
  booktitle={2021 IEEE/RSJ International Conference on Intelligent Robots and Systems (IROS)}, 
  title={CLINS: Continuous-Time Trajectory Estimation for LiDAR-Inertial System}, 
  year={2021},
  volume={},
  number={},
  pages={6657-6663},
  doi={10.1109/IROS51168.2021.9636676}}

@InProceedings{Behley19iccv_SemanticKITTI,
author = {Behley, Jens and Garbade, Martin and Milioto, Andres and Quenzel, Jan and Behnke, Sven and Stachniss, Cyrill and Gall, Jurgen},
title = {SemanticKITTI: A Dataset for Semantic Scene Understanding of LiDAR Sequences},
booktitle = {Proceedings of the IEEE/CVF International Conference on Computer Vision (ICCV)},
month = {October},
year = {2019}
}

@InProceedings{Caesar20cvpr_NuScenes,
author = {Caesar, Holger and Bankiti, Varun and Lang, Alex H. and Vora, Sourabh and Liong, Venice Erin and Xu, Qiang and Krishnan, Anush and Pan, Yu and Baldan, Giancarlo and Beijbom, Oscar},
title = {nuScenes: A Multimodal Dataset for Autonomous Driving},
booktitle = {Proceedings of the IEEE/CVF Conference on Computer Vision and Pattern Recognition (CVPR)},
month = {June},
year = {2020}
}

@inproceedings{cordtsCityscapesDatasetSemantic2016,
  title = {The {{Cityscapes Dataset}} for {{Semantic Urban Scene Understanding}}},
  booktitle = {Proceedings of the IEEE/CVF Conference on Computer Vision and Pattern Recognition (CVPR)},
  author = {Cordts, Marius and Omran, Mohamed and Ramos, Sebastian and Rehfeld, Timo and Enzweiler, Markus and Benenson, Rodrigo and Franke, Uwe and Roth, Stefan and Schiele, Bernt},
  year = {2016},
  pages = {3213--3223}
}

@inproceedings{Geiger12cvpr_KITTIAreWeReady,
  title = {Are We Ready for Autonomous Driving? {{The KITTI}} Vision Benchmark Suite},
  shorttitle = {Are We Ready for Autonomous Driving?},
  booktitle = {Proceedings of the IEEE/CVF Conference on Computer Vision and Pattern Recognition (CVPR)},
  author = {Geiger, Andreas and Lenz, Philip and Urtasun, Raquel},
  year = {2012},
  month = jun
}

@article{Geiger13ijrr_KITTIVisionMeetsRobotics,
  title = {Vision Meets Robotics: {{The KITTI}} Dataset},
  shorttitle = {Vision Meets Robotics},
  author = {Geiger, A and Lenz, P and Stiller, C and Urtasun, R},
  year = {2013},
  month = sep,
  journal = {The International Journal of Robotics Research},
  volume = {32},
  number = {11},
  pages = {1231--1237},
  publisher = {{SAGE Publications Ltd STM}},
  issn = {0278-3649},
  doi = {10.1177/0278364913491297},
  language = {en}
}

@InProceedings{Lang19cvpr_PointPillars,
    author = {Lang, Alex H. and Vora, Sourabh and Caesar, Holger and Zhou, Lubing and Yang, Jiong and Beijbom, Oscar},
    title = { {PointPillars}: Fast Encoders for Object Detection From Point Clouds},
    booktitle = {Proceedings of the IEEE/CVF Conference on Computer Vision and Pattern Recognition (CVPR)},
    month = {June},
    year = {2019}
}

@InProceedings{Liang20eccv_LaneGCN,
  title={Learning lane graph representations for motion forecasting},
  author={Liang, Ming and Yang, Bin and Hu, Rui and Chen, Yun and Liao, Renjie and Feng, Song and Urtasun, Raquel},
  booktitle = {ECCV},
  year={2020}
}

@inproceedings{Mercat20icra_MultiheadAttentionForecasting,
  title={Multi-head attention for multi-modal joint vehicle motion forecasting},
  author={Mercat, Jean and Gilles, Thomas and El Zoghby, Nicole and Sandou, Guillaume and Beauvois, Dominique and Gil, Guillermo Pita},
  booktitle={ICRA},
  year={2020},
  organization={IEEE}
}

@inproceedings{Sun20cvpr_WaymoOpenDataset,
  title = {Scalability in {{Perception}} for {{Autonomous Driving}}: {{Waymo Open Dataset}}},
  shorttitle = {Scalability in {{Perception}} for {{Autonomous Driving}}},
  booktitle = {Proceedings of the IEEE/CVF Conference on Computer Vision and Pattern Recognition (CVPR)},
  author = {Sun, Pei and Kretzschmar, Henrik and Dotiwalla, Xerxes and Chouard, Aurelien and Patnaik, Vijaysai and Tsui, Paul and Guo, James and Zhou, Yin and Chai, Yuning and Caine, Benjamin and Vasudevan, Vijay and Han, Wei and Ngiam, Jiquan and Zhao, Hang and Timofeev, Aleksei and Ettinger, Scott and Krivokon, Maxim and Gao, Amy and Joshi, Aditya and Zhang, Yu and Shlens, Jonathon and Chen, Zhifeng and Anguelov, Dragomir},
  year = {2020}
}

@InProceedings{Yin21cvpr_CenterPoint,
    author    = {Yin, Tianwei and Zhou, Xingyi and Krahenbuhl, Philipp},
    title     = {Center-Based 3D Object Detection and Tracking},
    booktitle = {Proceedings of the IEEE/CVF Conference on Computer Vision and Pattern Recognition (CVPR)},
    month     = {June},
    year      = {2021}
}

@article{ApolloScape,
  author    = {Yuexin Ma and
               Xinge Zhu and
               Sibo Zhang and
               Ruigang Yang and
               Wenping Wang and
               Dinesh Manocha},
  title     = {TrafficPredict: Trajectory Prediction for Heterogeneous Traffic-Agents},
  journal   = {CoRR},
  volume    = {abs/1811.02146},
  year      = {2018},
  archivePrefix = {arXiv},
  eprint    = {1811.02146},
}

@article{Liu22arxiv_BEVFusion,
  title={BEVFusion: Multi-Task Multi-Sensor Fusion with Unified Bird's-Eye View Representation},
  author={Liu, Zhijian and Tang, Haotian and Amini, Alexander and Yang, Xinyu and Mao, Huizi and Rus, Daniela and Han, Song},
  journal={arXiv preprint arXiv:2205.13542},
  year={2022}
}

@inproceedings{chang2019argoverse,
  title={Argoverse: 3d tracking and forecasting with rich maps},
  author={Chang, Ming-Fang and Lambert, John and Sangkloy, Patsorn and Singh, Jagjeet and Bak, Slawomir and Hartnett, Andrew and Wang, De and Carr, Peter and Lucey, Simon and Ramanan, Deva and others},
  booktitle={Proceedings of the IEEE/CVF conference on computer vision and pattern recognition},
  pages={8748--8757},
  year={2019}
}

@inproceedings{wu2020motionnet,
  title={Motionnet: Joint perception and motion prediction for autonomous driving based on bird's eye view maps},
  author={Wu, Pengxiang and Chen, Siheng and Metaxas, Dimitris N},
  booktitle={Proceedings of the IEEE/CVF conference on computer vision and pattern recognition},
  pages={11385--11395},
  year={2020}
}

@inproceedings{wang2022sti,
  title={Be-sti: Spatial-temporal integrated network for class-agnostic motion prediction with bidirectional enhancement},
  author={Wang, Yunlong and Pan, Hongyu and Zhu, Jun and Wu, Yu-Huan and Zhan, Xin and Jiang, Kun and Yang, Diange},
  booktitle={Proceedings of the IEEE/CVF Conference on Computer Vision and Pattern Recognition},
  pages={17093--17102},
  year={2022}
}

@inproceedings{li2022coda,
  title={Coda: A real-world road corner case dataset for object detection in autonomous driving},
  author={Li, Kaican and Chen, Kai and Wang, Haoyu and Hong, Lanqing and Ye, Chaoqiang and Han, Jianhua and Chen, Yukuai and Zhang, Wei and Xu, Chunjing and Yeung, Dit-Yan and others},
  booktitle={European Conference on Computer Vision},
  pages={406--423},
  year={2022},
  organization={Springer}
}

@inproceedings{xiao2021pandaset,
  title={Pandaset: Advanced sensor suite dataset for autonomous driving},
  author={Xiao, Pengchuan and Shao, Zhenlei and Hao, Steven and Zhang, Zishuo and Chai, Xiaolin and Jiao, Judy and Li, Zesong and Wu, Jian and Sun, Kai and Jiang, Kun and others},
  booktitle={2021 IEEE international intelligent transportation systems conference (ITSC)},
  pages={3095--3101},
  year={2021},
  organization={IEEE}
}

@inproceedings{dosovitskiy2017carla,
  title={CARLA: An open urban driving simulator},
  author={Dosovitskiy, Alexey and Ros, German and Codevilla, Felipe and Lopez, Antonio and Koltun, Vladlen},
  booktitle={Conference on robot learning},
  pages={1--16},
  year={2017},
  organization={PMLR}
}

@article{qian2025agentthink,
  title={Agentthink: A unified framework for tool-augmented chain-of-thought reasoning in vision-language models for autonomous driving},
  author={Qian, Kangan and Jiang, Sicong and Zhong, Yang and Luo, Ziang and Huang, Zilin and Zhu, Tianze and Jiang, Kun and Yang, Mengmeng and Fu, Zheng and Miao, Jinyu and others},
  journal={arXiv preprint arXiv:2505.15298},
  volume={1},
  number={2},
  pages={3},
  year={2025}
}

@misc{yu20264d,
      title={4D-ARE: Bridging the Attribution Gap in LLM Agent Requirements Engineering}, 
      author={Bo Yu and Lei Zhao},
      year={2026},
      eprint={2601.04556},
      archivePrefix={arXiv},
      primaryClass={cs.SE},
      url={https://arxiv.org/abs/2601.04556}, 
}

@inproceedings{qian2025lego,
  title={Lego-motion: Learning-enhanced grids with occupancy instance modeling for class-agnostic motion prediction},
  author={Qian, Kangan and Miao, Jinyu and Luo, Ziang and Fu, Zheng and Li, Jinchen and Shi, Yining and Wang, Yunlong and Jiang, Kun and Yang, Mengmeng and Yang, Diange},
  booktitle={2025 IEEE/RSJ International Conference on Intelligent Robots and Systems (IROS)},
  pages={14178--14185},
  year={2025},
  organization={IEEE}
}

@inproceedings{jiang2023vad,
  title={VAD: Vectorized scene representation for efficient autonomous driving},
  author={Jiang, Bo and Chen, Shaoyu and Xu, Qing and Liao, Bo and Chen, Jiayuan and Zhou, Hang and Zhang, Si and Liu, Chun and Huang, Caijing and Wang, Xiangwei and others},
  booktitle={Proceedings of the IEEE/CVF International Conference on Computer Vision (ICCV)},
  pages={8340--8350},
  year={2023}
}

@article{chen2024vad,
  title={VADv2: End-to-end vectorized autonomous driving via probabilistic planning},
  author={Chen, Shaoyu and Jiang, Bo and Gao, Hong and Liao, Bo and Xu, Qing and Zhang, Si and Huang, Caijing and Liu, Chun and Wang, Xiangwei},
  journal={arXiv preprint arXiv:2402.13243},
  year={2024}
}

@inproceedings{prakash2021transfuser,
  title={Multi-modal fusion transformer for end-to-end autonomous driving},
  author={Prakash, Aditya and Chitta, Kashyap and Geiger, Andreas},
  booktitle={Proceedings of the IEEE/CVF Conference on Computer Vision and Pattern Recognition},
  pages={7077--7087},
  year={2021}
}

@inproceedings{qian2025priormotion,
  title={Priormotion: Generative class-agnostic motion prediction with raster-vector motion field priors},
  author={Qian, Kangan and Miao, Jinyu and Jiao, Xinyu and Luo, Ziang and Fu, Zheng and Shi, Yining and Wang, Yunlong and Jiang, Kun and Yang, Diange},
  booktitle={Proceedings of the IEEE/CVF International Conference on Computer Vision},
  pages={27284--27294},
  year={2025}
}

@ARTICLE{thornton2021fmcw,
  author={Li, You and Ibanez-Guzman, Javier},
  journal={IEEE Signal Processing Magazine}, 
  title={Lidar for Autonomous Driving: The Principles, Challenges, and Trends for Automotive Lidar and Perception Systems}, 
  year={2020},
  volume={37},
  number={4},
  pages={50-61},
  doi={10.1109/MSP.2020.2973615}}

@INPROCEEDINGS{chen2023dicp, 
    AUTHOR    = {Bruno Hexsel AND Heethesh Vhavle AND Yi Chen}, 
    TITLE     = {{DICP: Doppler Iterative Closest Point Algorithm}}, 
    BOOKTITLE = {Proceedings of Robotics: Science and Systems}, 
    YEAR      = {2022}, 
    ADDRESS   = {New York City, NY, USA}, 
    MONTH     = {June}, 
    DOI       = {10.15607/RSS.2022.XVIII.015} 
}

@inproceedings{yoneda2023extended,
author = {Yubin Zeng and Youjin Yu and Shouzheng Qi and Tao Wu},
title = {Tracking 3D Moving Objects as Centroids Using FMCW LiDAR},
booktitle = {Proceedings of 4th 2024 International Conference on Autonomous Unmanned Systems (4th ICAUS 2024)},
editor = {Lianqing Liu and Yifeng Niu and Wenxing Fu and Yi Qu},
year = {2025},
publisher = {Springer Nature Singapore},
address = {Singapore},
pages = {536--545},
doi = {10.1007/978-981-96-3572-6_50},
isbn = {978-981-96-3572-6}
}

@ARTICLE{wu2022picking,
  author={Yoon, David J. and Chen, Yi and Vhavle, Heethesh and Reuther, James and Barfoot, Timothy D.},
  journal={IEEE Robotics and Automation Letters}, 
  title={Towards Fast Correspondence-Free Odometry Using Multiple FMCW Lidars}, 
  year={2025},
  volume={10},
  number={9},
  pages={9088-9095},
  doi={10.1109/LRA.2025.3592140}}

@InProceedings{yoon2023need,
    author    = {Chae, Yujeong and Park, Heejun and Kim, Hyeonseong and Yoon, Kuk-Jin},
    title     = {Doppler-Aware LiDAR-RADAR Fusion for Weather-Robust 3D Detection},
    booktitle = {Proceedings of the IEEE/CVF International Conference on Computer Vision (ICCV)},
    month     = {October},
    year      = {2025},
    pages     = {27197-27208}
}

@ARTICLE{zhao2024fmcw,
  author={Zhao, Mingle and Wang, Jiahao and Gao, Tianxiao and Xu, Chengzhong and Kong, Hui},
  journal={IEEE Robotics and Automation Letters}, 
  title={FMCW-LIO: A Doppler LiDAR-Inertial Odometry}, 
  year={2024},
  volume={9},
  number={6},
  pages={5727-5734},
  doi={10.1109/LRA.2024.3396636}}

@ARTICLE{zhao2024free,
  author={Rodríguez-Gómez, Juan Pablo and Tapia, Raul and Garcia, Maria del Mar Guzmán and Dios, Jose Ramiro Martínez-de and Ollero, Anibal},
  journal={IEEE Robotics and Automation Letters}, 
  title={Free as a Bird: Event-Based Dynamic Sense-and-Avoid for Ornithopter Robot Flight}, 
  year={2022},
  volume={7},
  number={2},
  pages={5413-5420},
  doi={10.1109/LRA.2022.3153904}}

@INPROCEEDINGS{gu2022iros,
  author={Gu, Yi and Cheng, Hongzhi and Wang, Kafeng and Dou, Dejing and Xu, Chengzhong and Kong, Hui},
  booktitle={2022 IEEE/RSJ International Conference on Intelligent Robots and Systems (IROS)}, 
  title={Learning Moving-Object Tracking with FMCW LiDAR}, 
  year={2022},
  volume={},
  number={},
  pages={3747-3753},
  doi={10.1109/IROS47612.2022.9981346}
  }

@INPROCEEDINGS{li2024perception,
  author={Cheng, Jen-Hao and Kuan, Sheng-Yao and Liu, Hou-I and Latapie, Hugo and Liu, Gaowen and Hwang, Jenq-Neng},
  booktitle={2024 IEEE International Conference on Image Processing (ICIP)}, 
  title={CenterRadarNet: Joint 3D Object Detection and Tracking Framework Using 4D FMCW Radar}, 
  year={2024},
  volume={},
  number={},
  pages={998-1004},
  doi={10.1109/ICIP51287.2024.10648077}}

@INPROCEEDINGS{sun2020scalability,
  author={Sun, Pei and Kretzschmar, Henrik and Dotiwalla, Xerxes and Chouard, Aurélien and Patnaik, Vijaysai and Tsui, Paul and Guo, James and Zhou, Yin and Chai, Yuning and Caine, Benjamin and Vasudevan, Vijay and Han, Wei and Ngiam, Jiquan and Zhao, Hang and Timofeev, Aleksei and Ettinger, Scott and Krivokon, Maxim and Gao, Amy and Joshi, Aditya and Zhang, Yu and Shlens, Jonathon and Chen, Zhifeng and Anguelov, Dragomir},
  booktitle={2020 IEEE/CVF Conference on Computer Vision and Pattern Recognition (CVPR)}, 
  title={Scalability in Perception for Autonomous Driving: Waymo Open Dataset}, 
  year={2020},
  volume={},
  number={},
  pages={2443-2451},
  doi={10.1109/CVPR42600.2020.00252}}

@article{jung2024helipr,
  title={HeLiPR: Heterogeneous LiDAR dataset for inter-LiDAR place recognition under spatiotemporal variations},
  author={Jung, Minwoo and Yang, Wooseong and Lee, Dongjae and Gil, Hyeonjae and Kim, Giseop and Kim, Ayoung},
  journal={The International Journal of Robotics Research},
  volume={43},
  number={12},
  pages={1867--1883},
  year={2024},
  publisher={SAGE Publications}
}

@inproceedings{shao2024lmdrive,
  title={Lmdrive: Closed-loop end-to-end driving with large language models},
  author={Shao, Hao and Hu, Yuxuan and Wang, Letian and Song, Guanglu and Waslander, Steven L and Liu, Yu and Li, Hongsheng},
  booktitle={Proceedings of the IEEE/CVF Conference on Computer Vision and Pattern Recognition},
  pages={15120--15130},
  year={2024}
}

@misc{huang2025ottervisionlanguageactionmodeltextaware,
      title={OTTER: A Vision-Language-Action Model with Text-Aware Visual Feature Extraction}, 
      author={Huang Huang and Fangchen Liu and Letian Fu and Tingfan Wu and Mustafa Mukadam and Jitendra Malik and Ken Goldberg and Pieter Abbeel},
      year={2025},
      eprint={2503.03734},
      archivePrefix={arXiv},
      primaryClass={cs.RO},
      url={https://arxiv.org/abs/2503.03734}, 
}

@article{mao2023gpt,
  title={Gpt-driver: Learning to drive with gpt},
  author={Mao, Jiageng and Qian, Yuxi and Ye, Junjie and Zhao, Hang and Wang, Yue},
  journal={arXiv preprint arXiv:2310.01415},
  year={2023}
}

@inproceedings{chen2024ppad,
  title={Ppad: Iterative interactions of prediction and planning for end-to-end autonomous driving},
  author={Chen, Zhili and Ye, Maosheng and Xu, Shuangjie and Cao, Tongyi and Chen, Qifeng},
  booktitle={European Conference on Computer Vision},
  pages={239--256},
  year={2024},
  organization={Springer}
}

@inproceedings{peebles2023scalable,
  title={Scalable diffusion models with transformers},
  author={Peebles, William and Xie, Saining},
  booktitle={Proceedings of the IEEE/CVF international conference on computer vision},
  pages={4195--4205},
  year={2023}
}

@article{chen2024end,
  title={End-to-end autonomous driving: Challenges and frontiers},
  author={Chen, Li and Wu, Penghao and Chitta, Kashyap and Jaeger, Bernhard and Geiger, Andreas and Li, Hongyang},
  journal={IEEE Transactions on Pattern Analysis and Machine Intelligence},
  year={2024},
  publisher={IEEE}
}

@inproceedings{weng2024paradrive,
  title={Para-drive: Parallelized architecture for real-time autonomous driving},
  author={Weng, Xinshuo and Ivanovic, Boris and Wang, Yan and Wang, Yue and Pavone, Marco},
  booktitle={Proceedings of the IEEE/CVF Conference on Computer Vision and Pattern Recognition},
  pages={15449--15458},
  year={2024}
}

@article{paden2016survey,
  title={A survey of motion planning and control techniques for self-driving urban vehicles},
  author={Paden, Brian and {\v{C}}{\'a}p, Michal and Yong, Sze Zheng and Yershov, Dmitry and Frazzoli, Emilio},
  journal={IEEE Transactions on intelligent vehicles},
  volume={1},
  number={1},
  pages={33--55},
  year={2016},
  publisher={IEEE}
}

@inproceedings{kendall2019learning,
  title={Learning to drive in a day},
  author={Kendall, Alex and Hawke, Jeffrey and Janz, David and Mazur, Przemyslaw and Reda, Daniele and Allen, John-Mark and Lam, Vinh-Dieu and Bewley, Alex and Shah, Amar},
  booktitle={2019 international conference on robotics and automation (ICRA)},
  pages={8248--8254},
  year={2019},
  organization={IEEE}
}

@article{bojarski2016end,
  title={End to end learning for self-driving cars},
  author={Bojarski, Mariusz and Del Testa, Davide and Dworakowski, Daniel and Firner, Bernhard and Flepp, Beat and Goyal, Prasoon and Jackel, Lawrence D and Monfort, Mathew and Muller, Urs and Zhang, Jiakai and others},
  journal={arXiv preprint arXiv:1604.07316},
  year={2016}
}

@article{touvron2023llama2,
  title={Llama 2: Open foundation and fine-tuned chat models},
  author={Touvron, Hugo and Martin, Louis and Stone, Kevin and Albert, Peter and Almahairi, Amjad and Babaei, Yasmine and Bashlykov, Nikolay and Batra, Soumya and Bhargava, Prajjwal and Bhosale, Shruti and others},
  journal={arXiv preprint arXiv:2307.09288},
  year={2023}
}

@article{touvron2023llama1,
  title={Llama: Open and efficient foundation language models},
  author={Touvron, Hugo and Lavril, Thibaut and Izacard, Gautier and Martinet, Xavier and Lachaux, Marie-Anne and Lacroix, Timoth{\'e}e and Rozi{\`e}re, Baptiste and Goyal, Naman and Hambro, Eric and Azhar, Faisal and others},
  journal={arXiv preprint arXiv:2302.13971},
  year={2023}
}

@article{wen2023dilu,
  title={Dilu: A knowledge-driven approach to autonomous driving with large language models},
  author={Wen, Licheng and Fu, Daocheng and Li, Xin and Cai, Xinyu and Ma, Tao and Cai, Pinlong and Dou, Min and Shi, Botian and He, Liang and Qiao, Yu},
  journal={arXiv preprint arXiv:2309.16292},
  year={2023}
}

@article{mao2023agentdriver,
  title={A language agent for autonomous driving},
  author={Mao, Jiageng and Ye, Junjie and Qian, Yuxi and Pavone, Marco and Wang, Yue},
  journal={arXiv preprint arXiv:2311.10813},
  year={2023}
}

@article{ishaq2025drivelmm,
  title={Drivelmm-o1: A step-by-step reasoning dataset and large multimodal model for driving scenario understanding},
  author={Ishaq, Ayesha and Lahoud, Jean and More, Ketan and Thawakar, Omkar and Thawkar, Ritesh and Dissanayake, Dinura and Ahsan, Noor and Li, Yuhao and Khan, Fahad Shahbaz and Cholakkal, Hisham and others},
  journal={arXiv preprint arXiv:2503.10621},
  year={2025}
}

@inproceedings{miyaoka2024chatmpc,
  title={Chatmpc: Natural language based mpc personalization},
  author={Miyaoka, Yuya and Inoue, Masaki and Nii, Tomotaka},
  booktitle={2024 American Control Conference (ACC)},
  pages={3598--3603},
  year={2024},
  organization={IEEE}
}

@inproceedings{altche2017lstm,
  title={An LSTM network for highway trajectory prediction},
  author={Altch{\'e}, Florent and de La Fortelle, Arnaud},
  booktitle={2017 IEEE 20th international conference on intelligent transportation systems (ITSC)},
  pages={353--359},
  year={2017},
  organization={IEEE}
}

@article{long2024vlmmpc,
  title={VLM-MPC: Vision Language Foundation Model (VLM)-Guided Model Predictive Controller (MPC) for Autonomous Driving},
  author={Long, Keke and Shi, Haotian and Liu, Jiaxi and Li, Xiaopeng},
  journal={arXiv preprint arXiv:2408.04821},
  year={2024}
}

@article{huang2024drivemm,
  title={Drivemm: All-in-one large multimodal model for autonomous driving},
  author={Huang, Zhijian and Feng, Chengjian and Yan, Feng and Xiao, Baihui and Jie, Zequn and Zhong, Yujie and Liang, Xiaodan and Ma, Lin},
  journal={arXiv preprint arXiv:2412.07689},
  year={2024}
}

@article{jiang2024senna,
  title={Senna: Bridging large vision-language models and end-to-end autonomous driving},
  author={Jiang, Bo and Chen, Shaoyu and Liao, Bencheng and Zhang, Xingyu and Yin, Wei and Zhang, Qian and Huang, Chang and Liu, Wenyu and Wang, Xinggang},
  journal={arXiv preprint arXiv:2410.22313},
  year={2024}
}

@article{zhang2024instruct,
  title={Instruct large language models to drive like humans},
  author={Zhang, Ruijun and Guo, Xianda and Zheng, Wenzhao and Zhang, Chenming and Keutzer, Kurt and Chen, Long},
  journal={arXiv preprint arXiv:2406.07296},
  year={2024}
}

@article{jiang2024koma,
  title={Koma: Knowledge-driven multi-agent framework for autonomous driving with large language models},
  author={Jiang, Kemou and Cai, Xuan and Cui, Zhiyong and Li, Aoyong and Ren, Yilong and Yu, Haiyang and Yang, Hao and Fu, Daocheng and Wen, Licheng and Cai, Pinlong},
  journal={IEEE Transactions on Intelligent Vehicles},
  year={2024},
  publisher={IEEE}
}

@article{wang2023drivemlm,
  title={Drivemlm: Aligning multi-modal large language models with behavioral planning states for autonomous driving},
  author={Wang, Wenhai and Xie, Jiangwei and Hu, ChuanYang and Zou, Haoming and Fan, Jianan and Tong, Wenwen and Wen, Yang and Wu, Silei and Deng, Hanming and Li, Zhiqi and others},
  journal={arXiv preprint arXiv:2312.09245},
  year={2023}
}

@article{hou2025driveagent,
  title={Driveagent: Multi-agent structured reasoning with llm and multimodal sensor fusion for autonomous driving},
  author={Hou, Xinmeng and Wang, Wuqi and Yang, Long and Lin, Hao and Feng, Jinglun and Min, Haigen and Zhao, Xiangmo},
  journal={arXiv preprint arXiv:2505.02123},
  year={2025}
}

@inproceedings{codevilla2018end,
  title={End-to-end driving via conditional imitation learning},
  author={Codevilla, Felipe and M{\"u}ller, Matthias and L{\'o}pez, Antonio and Koltun, Vladlen and Dosovitskiy, Alexey},
  booktitle={2018 IEEE international conference on robotics and automation (ICRA)},
  pages={4693--4700},
  year={2018},
  organization={IEEE}
}

@inproceedings{radford2021learning,
  title={Learning transferable visual models from natural language supervision},
  author={Radford, Alec and Kim, Jong Wook and Hallacy, Chris and Ramesh, Aditya and Goh, Gabriel and Agarwal, Sandhini and Sastry, Girish and Askell, Amanda and Mishkin, Pamela and Clark, Jack and others},
  booktitle={International conference on machine learning},
  pages={8748--8763},
  year={2021},
  organization={PmLR}
}

@article{sapkota2025vla,
  title={Vision-language-action models: Concepts, progress, applications and challenges},
  author={Sapkota, Ranjan and Cao, Yang and Roumeliotis, Konstantinos I and Karkee, Manoj},
  journal={arXiv preprint arXiv:2505.04769},
  year={2025}
}

@misc{zhou2025autovla,
      title={AutoVLA: A Vision-Language-Action Model for End-to-End Autonomous Driving with Adaptive Reasoning and Reinforcement Fine-Tuning}, 
      author={Zewei Zhou and Tianhui Cai and Seth Z. Zhao and Yun Zhang and Zhiyu Huang and Bolei Zhou and Jiaqi Ma},
      year={2025},
      eprint={2506.13757},
      archivePrefix={arXiv},
      primaryClass={cs.CV},
      url={https://arxiv.org/abs/2506.13757}, 
}

@inproceedings{liu2023bevfusion,
  title={Bevfusion: Multi-task multi-sensor fusion with unified bird's-eye view representation},
  author={Liu, Zhijian and Tang, Haotian and Amini, Alexander and Yang, Xinyu and Mao, Huizi and Rus, Daniela L and Han, Song},
  booktitle={2023 IEEE international conference on robotics and automation (ICRA)},
  pages={2774--2781},
  year={2023},
  organization={IEEE}
}

@article{crawshaw2020multi,
  title={Multi-task learning with deep neural networks: A survey},
  author={Crawshaw, Michael},
  journal={arXiv preprint arXiv:2009.09796},
  year={2020}
}

@inproceedings{zeng2019end,
  title={End-to-end interpretable neural motion planner},
  author={Zeng, Wenyuan and Luo, Wenjie and Suo, Simon and Sadat, Abbas and Yang, Bin and Casas, Sergio and Urtasun, Raquel},
  booktitle={Proceedings of the IEEE/CVF conference on computer vision and pattern recognition},
  pages={8660--8669},
  year={2019}
}

@article{naumann2025data,
  title={Data Scaling Laws for End-to-End Autonomous Driving},
  author={Naumann, Alexander and Gu, Xunjiang and Dimlioglu, Tolga and Bojarski, Mariusz and Degirmenci, Alperen and Popov, Alexander and Bisla, Devansh and Pavone, Marco and M{\"u}ller, Urs and Ivanovic, Boris},
  journal={arXiv preprint arXiv:2504.04338},
  year={2025}
}

@article{chitta2022transfuser,
  title={Transfuser: Imitation with transformer-based sensor fusion for autonomous driving},
  author={Chitta, Kashyap and Prakash, Aditya and Jaeger, Bernhard and Yu, Zehao and Renz, Katrin and Geiger, Andreas},
  journal={IEEE transactions on pattern analysis and machine intelligence},
  volume={45},
  number={11},
  pages={12878--12895},
  year={2022},
  publisher={IEEE}
}

@inproceedings{hu2023uniad,
  title={Planning-oriented autonomous driving},
  author={Hu, Yihan and Yang, Jiazhi and Chen, Li and Li, Keyu and Sima, Chonghao and Zhu, Xizhou and Chai, Siqi and Du, Senyao and Lin, Tianwei and Wang, Wenhai and others},
  booktitle={Proceedings of the IEEE/CVF conference on computer vision and pattern recognition},
  pages={17853--17862},
  year={2023}
}

@article{diffusiondrive,
  title={DiffusionDrive: Truncated Diffusion Model for End-to-End Autonomous Driving},
  author={Bencheng Liao and Shaoyu Chen and Haoran Yin and Bo Jiang and Cheng Wang and Sixu Yan and Xinbang Zhang and Xiangyu Li and Ying Zhang and Qian Zhang and Xinggang Wang},
  journal={arXiv preprint arXiv:2411.15139},
  year={2024}
}

@article{zheng2024preliminary,
  title={Preliminary Investigation into Data Scaling Laws for Imitation Learning-Based End-to-End Autonomous Driving},
  author={Zheng, Yupeng and Xia, Zhongpu and Zhang, Qichao and Zhang, Teng and Lu, Ben and Huo, Xiaochuang and Han, Chao and Li, Yixian and Yu, Mengjie and Jin, Bu and others},
  journal={arXiv preprint arXiv:2412.02689},
  year={2024}
}

@inproceedings{ishihara2021multi,
  title={Multi-task learning with attention for end-to-end autonomous driving},
  author={Ishihara, Keishi and Kanervisto, Anssi and Miura, Jun and Hautamaki, Ville},
  booktitle={Proceedings of the IEEE/CVF conference on computer vision and pattern recognition},
  pages={2902--2911},
  year={2021}
}

@inproceedings{sadat2020perceive,
  title={Perceive, predict, and plan: Safe motion planning through interpretable semantic representations},
  author={Sadat, Abbas and Casas, Sergio and Ren, Mengye and Wu, Xinyu and Dhawan, Pranaab and Urtasun, Raquel},
  booktitle={Computer Vision--ECCV 2020: 16th European Conference, Glasgow, UK, August 23--28, 2020, Proceedings, Part XXIII 16},
  pages={414--430},
  year={2020},
  organization={Springer}
}

@inproceedings{jaeger2023hidden,
  title={Hidden biases of end-to-end driving models},
  author={Jaeger, Bernhard and Chitta, Kashyap and Geiger, Andreas},
  booktitle={Proceedings of the IEEE/CVF International Conference on Computer Vision},
  pages={8240--8249},
  year={2023}
}

@article{ma2024surveyvla,
  title={A survey on vision-language-action models for embodied ai},
  author={Ma, Yueen and Song, Zixing and Zhuang, Yuzheng and Hao, Jianye and King, Irwin},
  journal={arXiv preprint arXiv:2405.14093},
  year={2024}
}

@inproceedings{xu2017end,
  title={End-to-end learning of driving models from large-scale video datasets},
  author={Xu, Huazhe and Gao, Yang and Yu, Fisher and Darrell, Trevor},
  booktitle={Proceedings of the IEEE conference on computer vision and pattern recognition},
  pages={2174--2182},
  year={2017}
}

@inproceedings{nuscenes2019,
  title={nuscenes: A multimodal dataset for autonomous driving},
  author={Caesar, Holger and Bankiti, Varun and Lang, Alex H and Vora, Sourabh and Liong, Venice Erin and Xu, Qiang and Krishnan, Anush and Pan, Yu and Baldan, Giancarlo and Beijbom, Oscar},
  booktitle={Proceedings of the IEEE/CVF conference on computer vision and pattern recognition},
  pages={11621--11631},
  year={2020}
}

@inproceedings{arai2025covla,
  title={Covla: Comprehensive vision-language-action dataset for autonomous driving},
  author={Arai, Hidehisa and Miwa, Keita and Sasaki, Kento and Watanabe, Kohei and Yamaguchi, Yu and Aoki, Shunsuke and Yamamoto, Issei},
  booktitle={2025 IEEE/CVF Winter Conference on Applications of Computer Vision (WACV)},
  pages={1933--1943},
  year={2025},
  organization={IEEE}
}

@inproceedings{waymo2024open,
  title={Scalability in perception for autonomous driving: Waymo open dataset},
  author={Sun, Pei and Kretzschmar, Henrik and Dotiwalla, Xerxes and Chouard, Aurelien and Patnaik, Vijaysai and Tsui, Paul and Guo, James and Zhou, Yin and Chai, Yuning and Caine, Benjamin and others},
  booktitle={Proceedings of the IEEE/CVF conference on computer vision and pattern recognition},
  pages={2446--2454},
  year={2020}
}

@article{wilson2023argoverse2,
  title={Argoverse 2: Next generation datasets for self-driving perception and forecasting},
  author={Wilson, Benjamin and Qi, William and Agarwal, Tanmay and Lambert, John and Singh, Jagjeet and Khandelwal, Siddhesh and Pan, Bowen and Kumar, Ratnesh and Hartnett, Andrew and Pontes, Jhony Kaesemodel and others},
  journal={arXiv preprint arXiv:2301.00493},
  year={2023}
}

@article{bartoccioni2025vavim,
  title={Vavim and vavam: Autonomous driving through video generative modeling},
  author={Bartoccioni, Florent and Ramzi, Elias and Besnier, Victor and Venkataramanan, Shashanka and Vu, Tuan-Hung and Xu, Yihong and Chambon, Loick and Gidaris, Spyros and Odabas, Serkan and Hurych, David and others},
  journal={arXiv preprint arXiv:2502.15672},
  year={2025}
}

@article{fu2025orion,
  title={Orion: A holistic end-to-end autonomous driving framework by vision-language instructed action generation},
  author={Fu, Haoyu and Zhang, Diankun and Zhao, Zongchuang and Cui, Jianfeng and Liang, Dingkang and Zhang, Chong and Zhang, Dingyuan and Xie, Hongwei and Wang, Bing and Bai, Xiang},
  journal={arXiv preprint arXiv:2503.19755},
  year={2025}
}

@article{rafailov2023direct,
  title={Direct preference optimization: Your language model is secretly a reward model},
  author={Rafailov, Rafael and Sharma, Archit and Mitchell, Eric and Manning, Christopher D and Ermon, Stefano and Finn, Chelsea},
  journal={Advances in Neural Information Processing Systems},
  volume={36},
  pages={53728--53741},
  year={2023}
}

@article{schulman2017proximal,
  title={Proximal policy optimization algorithms},
  author={Schulman, John and Wolski, Filip and Dhariwal, Prafulla and Radford, Alec and Klimov, Oleg},
  journal={arXiv preprint arXiv:1707.06347},
  year={2017}
}


\vfill

\end{document}